%% file: zhou.tex
\documentclass[letterpaper]{article} 
\usepackage{aaai2026}  
\copyrighttext{}
\usepackage{times}  
\usepackage{helvet}  
\usepackage{courier}  
\usepackage[hyphens]{url}  
\usepackage{graphicx} 
\usepackage{natbib}  
\usepackage{caption} 
\usepackage{algorithm}
\usepackage{algorithmic}

\usepackage{amsmath}
\usepackage{amssymb}
\usepackage{mathtools}
\usepackage{amsthm}
\usepackage[capitalize,noabbrev]{cleveref}
\usepackage{booktabs}
\usepackage{multirow}
\usepackage{csquotes}

\usepackage{newfloat}
\usepackage{listings}
\DeclareCaptionStyle{ruled}{labelfont=normalfont,labelsep=colon,strut=off} 
\floatstyle{ruled}
\newfloat{listing}{tb}{lst}{}
\floatname{listing}{Listing}
\title{Dynamic Weight Adaptation in Spiking Neural Networks \\ Inspired by Biological Homeostasis\footnote{Accepted at AAAI 2026. This is the extended version including technical appendices.}}
\author{
    Yunduo Zhou\textsuperscript{\rm 1}, 
    Bo Dong\textsuperscript{\rm 2}, 
    Chang Li\textsuperscript{\rm 1}, 
    Yuanchen Wang\textsuperscript{\rm 1}, 
    Xuefeng Yin\textsuperscript{\rm 1}, 
    Yang Wang\textsuperscript{\rm 1}, 
    Xin Yang\textsuperscript{\rm 1}\thanks{Corresponding author.}
}
\affiliations{
    \textsuperscript{\rm 1}Key Laboratory of Social Computing and Cognitive Intelligence, Dalian University of Technology\\
    \textsuperscript{\rm 2}Cephia AI

    zyd\_dlut@163.com, bo.dong@cephia.ai, lchang@mail.dlut.edu.cn, , \\xfyin@mail.dlut.edu.cn, yangwang06@mail.dlut.edu.cn, xinyang@dlut.edu.cn
}

\usepackage{bibentry}

\begin{document}

\maketitle

\begin{abstract}
Homeostatic mechanisms play a crucial role in maintaining optimal functionality within the neural circuits of the brain. By regulating physiological and biochemical processes, these mechanisms ensure the stability of an organism’s internal environment, enabling it to better adapt to external changes. Among these mechanisms, the Bienenstock, Cooper, and Munro (BCM) theory has been extensively studied as a key principle for maintaining the balance of synaptic strengths in biological systems. Despite the extensive development of spiking neural networks (SNNs) as a model for bionic neural networks, no prior work in the machine learning community has integrated biologically plausible BCM formulations into SNNs to provide homeostasis. In this study, we propose a Dynamic Weight Adaptation Mechanism (DWAM) for SNNs, inspired by the BCM theory. DWAM can be integrated into the host SNN, dynamically adjusting network weights in real time to regulate neuronal activity, providing homeostasis to the host SNN without any fine-tuning. We validated our method through dynamic obstacle avoidance and continuous control tasks under both normal and specifically designed degraded conditions. Experimental results demonstrate that DWAM not only enhances the performance of SNNs without existing homeostatic mechanisms under various degraded conditions but also further improves the performance of SNNs that already incorporate homeostatic mechanisms.
\end{abstract}


\section{Introduction}

Homeostasis in biological organisms is a critical mechanism for maintaining survival and functionality by regulating physiological and biochemical processes to ensure internal environmental stability~\cite{marder2006variability,  desai2003homeostatic, turrigiano2004homeostatic, marder2002modeling, mease2013emergence}. In the nervous system, homeostasis can be achieved through various mechanisms such as synaptic scaling, excitatory-inhibitory balance, Bienenstock, Cooper, and Munro (BCM) theory, and dynamic firing thresholds~\cite{desai2003homeostatic, keck2017interactions, turrigiano2004homeostatic}. These mechanisms finely tune and balance neuronal activity, ensuring that the nervous system maintains a relatively stable activity level in response to external stimuli or internal disturbances, thereby preventing excessive excitation or inhibition of neurons. Several studies have shown that the absence or dysregulation of neuronal homeostasis can cause neuronal activity levels to escalate rapidly, resulting in network dysfunction ~\cite{o2014cell, marder2002modeling}.

Given the critical role of homeostasis in biological neural networks, many researchers have sought to incorporate this mechanism into spiking neural networks (SNNs) to enhance networks’ performance~\cite{meng2011modeling, hao2020biologically, kim2021spiking, xu2022endowing, sun2021predicting}. Due to the bio-inspired nature of SNNs, incorporating biological mechanisms into them is relatively straightforward~\cite{roy2019towards, maass1997networks}, making SNNs an ideal platform for such endeavors. However, while previous attempts have introduced homeostatic mechanisms into SNNs, most of these approaches have been heuristic or empirical, lacking biological interpretability and failing to validate the effectiveness of neural homeostasis in artificial systems. Ding  et al.~\cite{ding2022biologically} addressed this gap by proposing a biologically plausible method to implement homeostasis in SNNs through dynamic firing thresholds. Inspired by the adaptive thresholding observed in barn owl cochlear neurons, their method calculates the firing threshold of each neuron in real time based on membrane potential, dynamically modulating neuronal activity levels. This approach successfully introduces homeostasis into SNNs by regulating intrinsic neuronal properties.

Nonetheless, achieving homeostasis solely by adjusting intrinsic neuronal properties is incomplete. In biological systems, homeostasis depends not only on the regulation of individual neurons but also on the dynamic balance of inter-neuronal relationships~\cite{desai2003homeostatic, keck2017interactions}. The BCM theory~\cite{bienenstock1982theory}, a classical model of synaptic plasticity, is a key mechanism for maintaining homeostasis by adapting synaptic weights based on neurons’ historical activity levels to stabilize subsequent neuronal responses. Based on this theory, we developed the Dynamic Weight Adaptation Mechanism (DWAM), a mechanism that dynamically adjusts network weights in real-time according to the firing relationships and activity levels of neurons. Specifically, DWAM provides two key contributions: 1)	It assesses firing relationships between neurons to evaluate their correlation, enhancing weights between highly correlated neurons while weakening those between weakly correlated ones, thereby maintaining accurate inter-neuronal associations. 2) It dynamically determines the direction of weight modifications based on neuronal activity levels, preventing sustained enhancement of highly correlated neurons or persistent weakening of weakly correlated neurons, ensuring long-term balance and network stability. 

While DWAM is inspired by the BCM theory, maintaining consistency with its principles and formulaic structure, the inherent differences between SNNs and biological neural systems make the direct application of the original biological BCM model ineffective for introducing homeostasis into SNNs. Therefore, we analyze the original BCM model and leverage the statistical firing patterns of SNN neurons to refine the strength of weight adjustments, tailoring it for application in SNNs.

We validated our method through dynamic obstacle avoidance and continuous control tasks under both normal and specifically designed degraded conditions. Across all scenarios, the developed BCM mechanism consistently delivered performance improvements. Notably, these gains were achieved by directly integrating DWAM into the host SNNs, without any fine-tuning.

In particular, we make the following contributions in this work:
\begin{itemize}
\item We propose a dynamic weight adaptation mechanism for SNNs, which represents the first implementation of biologically inspired homeostasis from the perspective of inter-neuronal relationships, based on the BCM theory.

\item We design an approach that optimizes the original biological model’s weight adjustment strength using statistical firing cues from neurons, ensuring higher stability and feasibility in SNNs.

\item We validate the effectiveness of the proposed method across multiple tasks and degraded conditions, demonstrating that DWAM achieves biologically inspired homeostasis and significantly enhances SNN generalization across diverse tasks and conditions. Notably, all results are obtained by directly integrating DWAM into host SNNs without any fine-tuning, highlighting its plug-and-play nature.
\end{itemize}

\section{Background and Related Work}
\subsection{Homeostasis and BCM Theory}
\label{BCM}
The BCM theory provides a framework for understanding synaptic plasticity by describing how synaptic weights change in response to neuronal activity. According to BCM theory, the synaptic modification is governed by the rule~\cite{bienenstock1982theory}:
\begin{align}
\frac{dw_{ij}}{dt} &= \phi(x_{i}) x_{j} \label{eq:bio_w}\\
\phi(x_{i})&=x_{i}(x_{i}-\theta _{M}) \label{eq:bio_phi}
\end{align}
where $w_{ij}$ is the synaptic weight of the connection between neuron $i$ and neuron $j$, $x_{j}$ is the presynaptic activity, $x_{i}$ is the postsynaptic activity, and $\phi(x_{i})$ is a function of the postsynaptic activity. Here, $\theta _{M}$ is a sliding modification threshold that adapts based on the history of postsynaptic activity and ensures a balance between synaptic potentiation (when $x_{i}>\theta _{M}$) and depression (when $x_{i}<\theta _{M}$). The sliding modification threshold $\theta _{M}$ is typically given by:
\begin{align}
\theta _{M} = \left \langle x_{i}^{2} \right \rangle \label{eq:bio_theta}
\end{align}
where $\left \langle x_{i}^{2} \right \rangle $ represents the long-term average of the squared postsynaptic activity. 

\subsection{Spiking Neural Networks (SNNs)}

SNNs are a class of biologically inspired neural networks that aim to replicate the information processing mechanisms of biological neurons. Unlike traditional networks that process continuous signals, SNNs transmit information through discrete spikes, which occur when a neuron’s membrane potential exceeds a certain threshold. These spikes are generated by the integration of incoming signals over time. When the potential reaches the threshold, the neuron “fires”, sending a spike to neighboring neurons, and the potential resets.

Over the years, various neuron models have been developed to simulate the dynamics of spiking neurons, with two commonly used ones being the Spike Response Model (SRM)~\cite{gerstner1995time} and the Leaky Integrate-and-Fire (LIF) model~\cite{gerstner2002spiking}. More details about these two models are provided in Supplementary Note 2. 

\subsection{SNN with Homeostasis}
Although homeostatic mechanisms are well studied in biology, their integration in machine learning is limited. Some works adopt BCM or STDP to guide SNN learning~\cite{meng2011modeling, pan2023adaptive, kheradpisheh2018stdp, gupta2024unsupervised, wang2024apprenticeship}, but they are task-specific or applied only during training, offering no functional homeostasis during inference and thus not directly comparable to our approach.

Other works have attempted to introduce homeostatic effects during the testing phase by adapting neuronal parameters such as thresholds or synaptic strengths~\cite{hao2020biologically, kim2021spiking, xu2022endowing, zhang2025spiking, zhang2022spiking, wang2023event}. However, these approaches are largely empirical or heuristic in nature, lacking biological interpretability and principled justification. 

In contrast, Ding  et al.~\cite{ding2022biologically} proposed a biologically inspired method for calculating the dynamic membrane potential threshold in SNNs. This threshold is related to the neuron's membrane potential and the previous depolarization rate. They also demonstrated the advantages of their method compared to the aforementioned approaches that lack biological grounding. Therefore, the primary comparison method in this study is that of Ding et al.

\section{Method}

\begin{figure*}[ht]
\begin{center}
\centerline{\includegraphics[width=14cm]{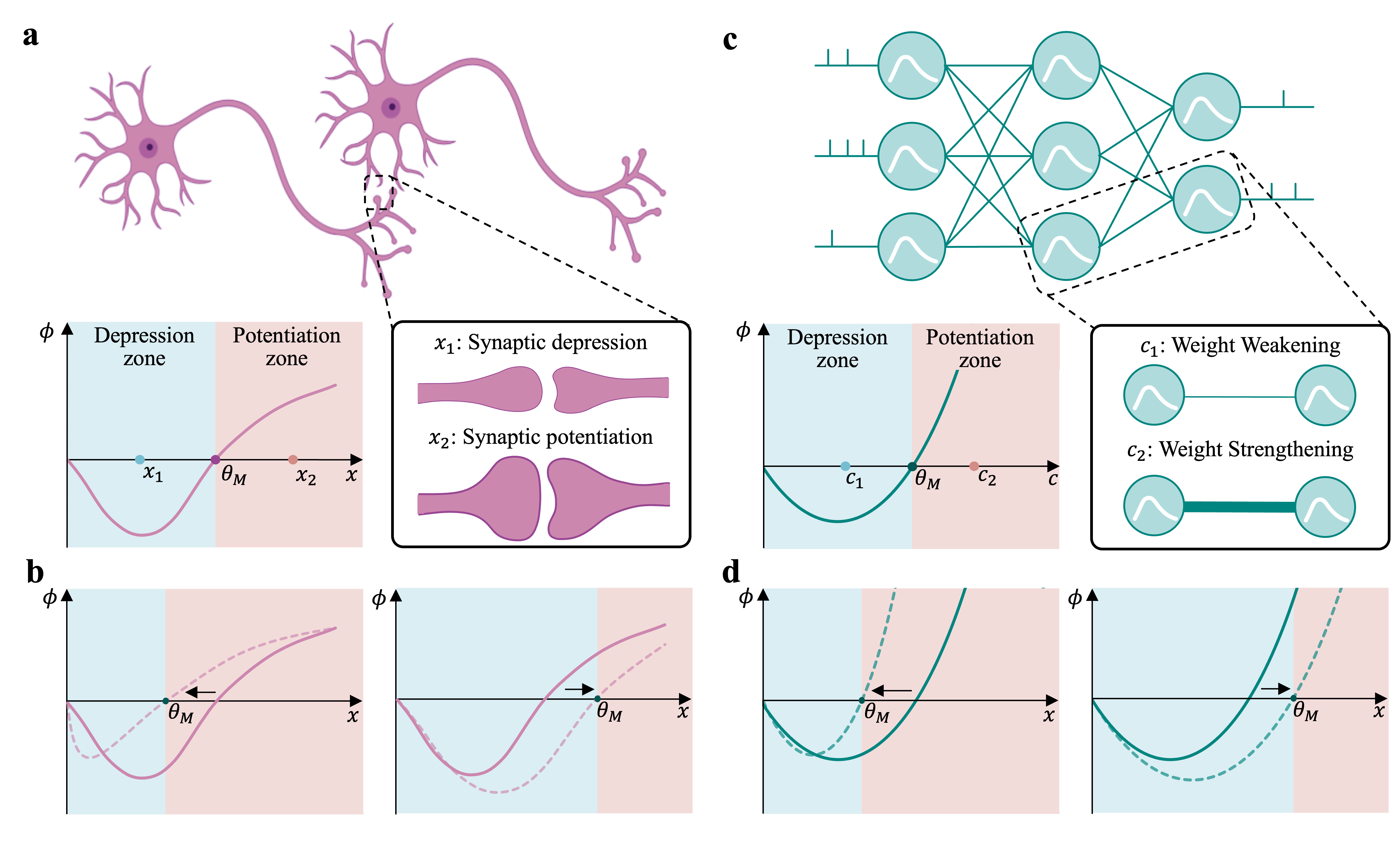}}
\caption{\textbf{Comparison between biological BCM theory and the proposed DWAM in SNNs.} a. Illustration of the BCM theory in biological neural systems. The synaptic modification function ($\phi$) is governed by the postsynaptic neuron’s activity ($x$) and a modification threshold ($\theta_M$). The depression zone ($x < \theta_M$) leads to synaptic weakening (e.g., $x_1$), while the potentiation zone ($x > \theta_M$) results in synaptic strengthening (e.g., $x_2$). b. The dynamic adjustment of the modification threshold ($\theta_M$) based on the historical activity level of the postsynaptic neuron. A higher historical activity level shifts $\theta_M$ to the right, reducing potentiation likelihood, whereas lower historical activity shifts $\theta_M$ to the left, increasing potentiation likelihood.(illustrated as a shift from the solid line to the dashed line in the figure) c. Proposed DWAM for SNNs, inspired by the BCM theory. The synaptic modification function ($\phi$) in DWAM operates on a similar principle, where the postsynaptic neuron’s firing rate ($c$) interacts with a modification threshold ($\theta_M$) to determine weight weakening ($c_1$) or strengthening ($c_2$). d. As with BCM, a higher historical firing rate shifts $\theta_M$ to the right, whereas a lower firing rate shifts it to the left, enabling balanced and stable weight updates in the SNNs.}
\label{fig:overview}
\end{center}
\end{figure*}

The core concept of the biological BCM theory can be divided into two main components, which describe the mechanism of synaptic strength adjustment and the mechanism for maintaining stability.

The first component describes how synaptic strength is adjusted based on the postsynaptic neuron’s activity relative to a modification threshold. As shown in Figure~\ref{fig:overview}a, when the presynaptic neuron fires and the postsynaptic activity (e.g., $x_1$) is below the threshold ($\theta_M$), the synapse is considered weakly correlated and is suppressed. Conversely, when the activity (e.g., $x_2$) exceeds $\theta_M$, the synapse is strengthened. This mechanism dynamically modulates synaptic strength according to neuronal correlation, promoting balanced synaptic regulation.

The second component governs the dynamic adaptation of the modification threshold itself based on the postsynaptic neuron’s activity history (Figure~\ref{fig:overview}b). Low historical activity lowers $\theta_M$, increasing the chance of synaptic strengthening, while high activity raises $\theta_M$, promoting synaptic weakening. This prevents unbounded potentiation or depression, maintaining synaptic stability over time.

In this study, we incorporate the BCM theory into SNNs to achieve a biologically inspired homeostatic mechanism. While prior work such as BDETT~\cite{ding2022biologically} introduces homeostasis by regulating intrinsic neuronal properties during training, our approach focuses on inter-neuronal relationships, providing a complementary and orthogonal mechanism that can coexist with intrinsic regulation. Importantly, unlike BDETT which requires retraining to exploit noise resilience, our method is plug-and-play and can be applied directly to pre-trained SNNs during inference, enabling homeostasis without extra computational cost. This combination of biological plausibility, orthogonal homeostatic strategy, and deployment efficiency makes our approach both practically and conceptually significant.

\cref{sec:SNN_BCM_1} elaborates on how to implement the synaptic strength adjustment mechanism in SNNs, corresponding to the first component of the BCM theory. \cref{sec:SNN_BCM_2} introduces the realization of the stability maintenance mechanism in SNNs, addressing related challenges and presenting our solutions. Together, these two components form the proposed DWAM. 

\subsection{The Synaptic Adjustment Mechanism}
\label{sec:SNN_BCM_1}
The synaptic strength adjustment mechanism in SNNs can be implemented by substituting the firing rate $c$ of spiking neurons for the activity level $x$ of biological neurons in Eqs.~\ref{eq:bio_w} and ~\ref{eq:bio_phi}. At a given timestamp $t$, the modification of the weight from the $j$-th neuron in the $l-1$-th layer to the $i$-th neuron in the $l$-th layer is defined as follows:
\begin{align}
w_{ij}(t) &= w_{ij}(t{-}1) + \phi_{ij}(t) c_{j}^{l{-}1}(t) |w_{ij}(t{-}1)| \psi, \label{eq:w} \\
\phi_{ij}(t) &= c_{i}^{l}(t) \left(c_{i}^{l}(t) - \theta_{i}^{l}(t)\right) \label{eq:phi}
\end{align}
where $\phi$ is the modification function for the weights; $c$ is the firing rate of the neuron; $\theta$ is the sliding modification threshold; $w_{ij}$ is the weight from the $j$-th neuron in the $l-1$-th layer to the $i$-th neuron in the $l$-th layer, and $\psi$ is a constant. In this implementation, we employ  $|w|\psi$ as a scaling factor to dynamically adjust the magnitude of weight changes based on the current weight value. This design ensures that the adjustment for smaller weights does not become excessively large, preventing undue disruption to their stability. Similarly, it allows larger weights to receive sufficient adjustments, avoiding under-scaling that could reduce the effectiveness of weight modulation. 

According to these formulas, the weight change is proportional to the product of the modification function and the presynaptic neuron firing rate. Additionally, the postsynaptic neuron firing rate is governed by a dynamic threshold level ( $\theta$ ). When the current firing rate exceeds this threshold, the weight is strengthened; conversely, when the firing rate falls below the threshold, the weight is weakened. As shown in Figure~\ref{fig:overview}c, this process aligns with the first component of the biological BCM theory.

\subsection{The Stability Maintenance Mechanism}
\label{sec:SNN_BCM_2}

According to the BCM theory, the sliding modification threshold $\theta_M$ in Eq.\ref{eq:bio_theta} is a super-linear function of the history of postsynaptic activity. Benuskova et al.~\cite{benuskova1994dynamic} developed a computational model of a single representative barrel cell based on the BCM theory, where $\theta_M$ is defined as:

\begin{align}
\theta_M (t) &= \left [ \frac{\left \langle c^{2}(t)  \right \rangle_{\tau  } }{\eta}  \right ], \label{eq:barrel_theta} \\
\left \langle c^{2}(t)  \right \rangle_{\tau  } &=\frac{1}{\tau} \int_{-\infty}^{t} c^{2}(t^{'}) e^{-\frac{(t-t^{'})}{\tau}}dt^{'} \label{eq:barrel_c}
\end{align}
where $\eta $ and $\tau$ are positive scaling constants. From these relations, it follows that $\theta_M (t)$ is positively correlated with the past firing rate, and thus it can be used to measure the averaged neuron activity in the recent past. To make it applicable to SNNs, we discretize Eqs.~\ref{eq:barrel_theta} and ~\ref{eq:barrel_c} as:
\begin{align}
\theta_{M,i}^{l}(t) = (1 - \alpha) \theta_{M,i}^{l}(t-1) + \alpha ( c_{i}^{l} (t))^{2} \label{eq:SNN_theta_bio}
\end{align}
This is a commonly used recursive form of exponential moving average, where $\alpha$ controls the update rate and corresponds to the decay constant of the sliding modification threshold.

Although Eqs.~\ref{eq:w}, ~\ref{eq:phi}, and~\ref{eq:SNN_theta_bio} successfully reproduce the BCM theory in biological neural systems, when $\theta_M$ defined in Eq.~\ref{eq:SNN_theta_bio} is used as $\theta_{i}^{l} (t)$ in Eq.~\ref{eq:phi}, the resulting SNN applying Eqs.~\ref{eq:w}, ~\ref{eq:phi}, and~\ref{eq:SNN_theta_bio} fails to achieve the expected performance. To distinguish this approach from the final method, we refer to the method based on these three equations as Biological-Originated DWAM (BioDWAM). This discrepancy arises primarily due to differences in neuronal firing rates. In biological systems, neurons typically maintain relatively low firing rates to ensure sparse coding and energy efficiency~\cite{olshausen1996emergence, stringer2019high}. However, in SNNs, some neurons may need to sustain high firing rates for extended periods to meet task requirements and maintain network performance. When a neuron exhibits prolonged high firing rates, the sliding modification threshold gradually increases and stabilizes at a high level. According to Eqs.~\ref{eq:w} and ~\ref{eq:phi}, this leads to continuous weakening of the weights connected to the neuron, preventing the network from reaching a stable state.

\begin{figure}[htbp]
\begin{center}
\centerline{\includegraphics[width=8cm]{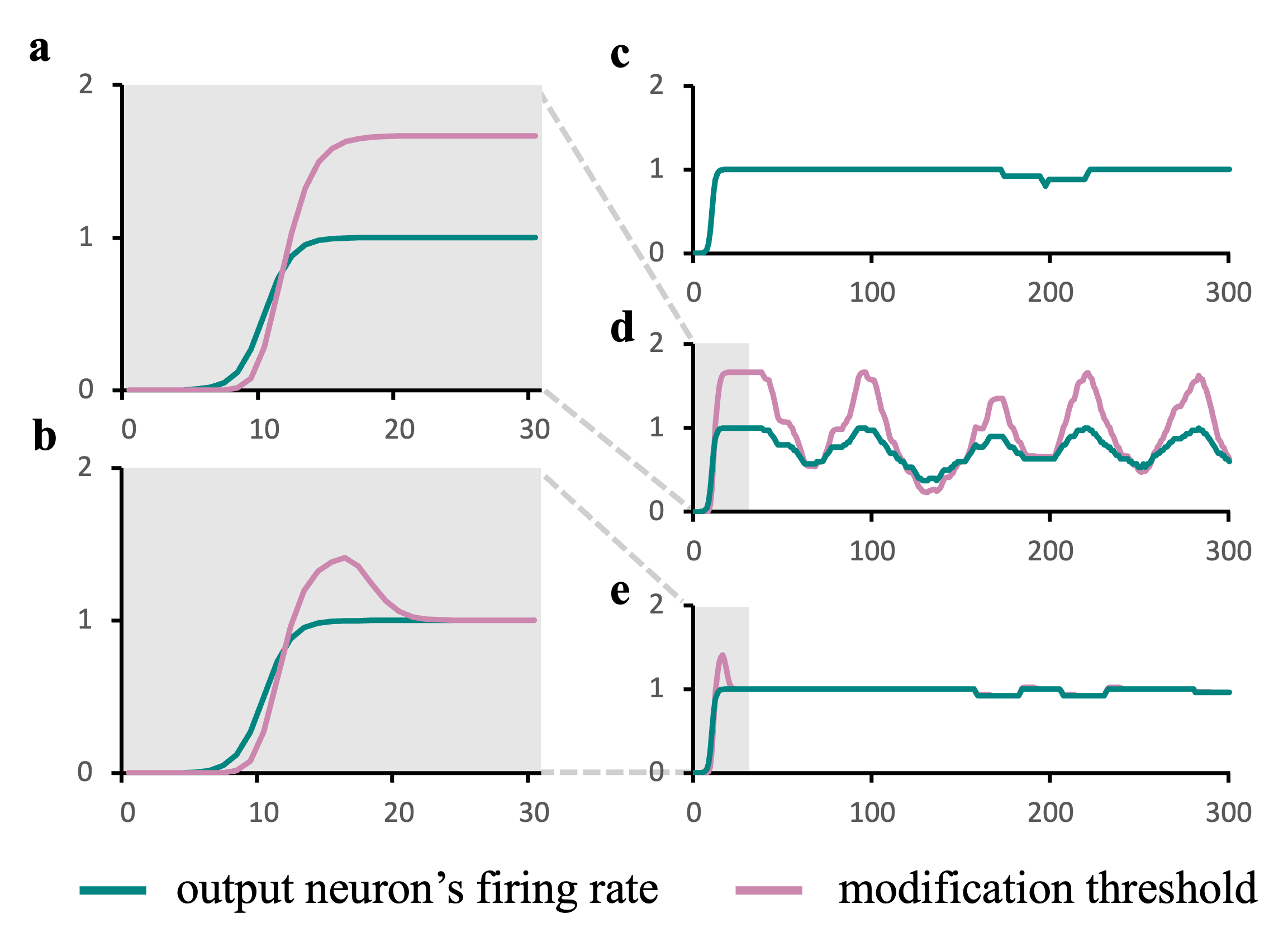}}
\caption{\textbf{Behavioral comparison between BioDWAM and DWAM in the toy example.} 
All subplots show simulation steps (x-axis) versus value (y-axis), with green and pink curves indicating firing rate and modification threshold, respectively.
a \& b. Zoomed-in views of the initial phase in d and e, showing early threshold adaptation. c. Firing rate stabilizes at 1 without synaptic adjustment. d. BioDWAM induces oscillatory instability at high firing rates. e. DWAM maintains stability by modulating threshold according to firing variability.}
\label{fig:toyexample}
\end{center}
\end{figure}

To illustrate this phenomenon, we designed a toy example. In this setup, the output layer of the network contains only a single neuron, trained to maintain a high firing rate, ideally close to 1. We applied BioDWAM directly to the output layer of this network and recorded the changes in the output neuron’s firing rate and modification threshold over time. See Supplementary Note 3 for further training details.

Figure~\ref{fig:toyexample}c shows the firing rate of the output neuron without the application of BioDWAM. The firing rate gradually increases and stabilizes at 1, representing the ideal behavior of the network. In contrast, Figure ~\ref{fig:toyexample}d illustrates the behavior when BioDWAM is applied. Initially, when the output neuron’s firing rate is low, the sliding modification threshold remains within a reasonable range. However, as the firing rate increases to a high level, the threshold rises significantly. According to Eqs.~\ref{eq:w} and ~\ref{eq:phi}, this results in a weakening of the weights connected to the neuron. As a consequence, the firing rate begins to decrease. Once the firing rate falls below the sliding modification threshold, the weights are strengthened again, causing the firing rate to rise once more. This cycle continues, leading to an unstable oscillatory output characterized by wave-like patterns.

This observation highlights that directly applying the original biological synaptic adjustment model to SNNs can introduce stability issues, particularly when dealing with neurons with high firing rates. 

\textbf{Our Solution} 
To retain the original stabilizing effect of the BCM theory in neural networks and to avoid the instability caused by high firing-rate neurons, we introduce a balancing term in Eq.~\ref{eq:SNN_theta_bio}:
\begin{align}
\theta_{i}^{l}(t) &= \theta_{M,i}^{l}(t) \zeta \frac{\sigma (c_{i}^{l},n)}{\mu (c_{i}^{l},n)} + c_{i}^{l} (t) (1-\zeta \frac{\sigma (c_{i}^{l},n)}{\mu (c_{i}^{l},n)}) \label{eq:SNN_theta}
\end{align}
Here, $\zeta$ is a positive scaling constant, and $\sigma$ and $\mu$ denote the standard deviation and mean of the postsynaptic neuron’s firing rates over the past $n$ time steps. Their ratio, $\frac{\sigma}{\mu}$, is the coefficient of variation (CV), which reflects the stability of neuronal activity.

As shown in Eq.\ref{eq:SNN_theta}, the CV modulates the influence of the biologically originated sliding threshold $\theta_{M,i}^{l}(t)$ on $\theta_{i}^{l}(t)$. A high CV (i.e., unstable firing) increases the influence of $\theta_{M,i}^{l}(t)$, pulling $\theta_{i}^{l}(t)$ toward it. Conversely, a low CV (i.e., stable firing) causes $\theta_{i}^{l}(t)$ to align more closely with the current firing rate. As indicated in Eqs.\ref{eq:w} and \ref{eq:phi}, when $\theta_{i}^{l}(t)$ matches the firing rate, weight updates diminish, preventing persistent high-frequency firing from continuously weakening synaptic weights.

We applied DWAM to the same toy example as before. As shown in Figure~\ref{fig:toyexample}e, DWAM effectively mitigates the instability observed in BioDWAM at high firing rates. To better illustrate the behavioral differences, the early phases of Figure~\ref{fig:toyexample}d and e are enlarged in Figure~\ref{fig:toyexample}a and b. Initially, DWAM behaves similarly to BioDWAM, but once the neuron stabilizes at a high firing rate, DWAM’s correction threshold gradually converges to that rate. This allows DWAM to retain the biologically consistent behavior of the BCM theory at low firing rates while avoiding the instability seen when directly applying the biological model to SNNs.

By integrating Eqs.~\ref{eq:w}, ~\ref{eq:phi}, and ~\ref{eq:SNN_theta}, we successfully incorporate the BCM theory into SNNs, providing biologically-plausible homeostasis for the network.

\section{Experiments and Results}

\begin{table*}[t]
\centering
\caption{Quantitative performance of the robotic obstacle avoidance task under degraded conditions, where higher SR indicates better performance, and lower $\text{HM}_{m}$ and $\text{HM}_{std}$ values reflect improved homeostasis.} 
\label{tab:OA-Degraded}
\begin{small}
\resizebox{17.5cm}{!}{
\setlength\tabcolsep{1pt}
\begin{tabular}{ccccccccccccc} 
\toprule
                       & \multicolumn{4}{c}{\textbf{0.2 }}                                                                           & \multicolumn{4}{c}{\textbf{6.0}}                                                                            & \multicolumn{4}{c}{\textbf{GN}}                                                                              \\ 
\cmidrule(lr){2-5}\cmidrule(lr){6-9}\cmidrule(lr){10-13}
                       & \multicolumn{2}{c}{LIF}                              & \multicolumn{2}{c}{SRM}                              & \multicolumn{2}{c}{LIF}                              & \multicolumn{2}{c}{SRM}                              & \multicolumn{2}{c}{LIF}                              & \multicolumn{2}{c}{SRM}                               \\ 
\cmidrule(lr){2-3}\cmidrule(lr){4-5}\cmidrule(lr){6-7}\cmidrule(lr){8-9}\cmidrule(lr){10-11}\cmidrule(lr){12-13}
\textbf{\textbf{Name}} & SR                   & $\text{HM}_m$/$\text{HM}_{std}$ & SR                   & $\text{HM}_m$/$\text{HM}_{std}$ & SR                   & $\text{HM}_m$/$\text{HM}_{std}$ & SR                   & $\text{HM}_m$/$\text{HM}_{std}$ & SR                   & $\text{HM}_m$/$\text{HM}_{std}$ & SR                   & $\text{HM}_m$/$\text{HM}_{std}$  \\ 
\midrule
SAN                 & 88.2\%               & 0.109/0.115                   & 86.0\%               & 0.130/0.117                   & 71.7\%               & \textbf{0.018/0.022}          & 75.6\%               & 0.027/0.028                   & 87.6\%               & 0.033/0.031                   & 81.6\%               & 0.039/\textbf{0.034}           \\
SAN-DWAM            & \textbf{90.6\%}      & \textbf{0.091/0.113}          & \textbf{87.3\%}      & \textbf{0.121/0.112}          & \textbf{84.3\%}      & \textbf{0.018/0.022}          & \textbf{76.1\%}      & \textbf{0.026/0.027}          & \textbf{88.4\%}      & \textbf{0.031/0.030}          & \textbf{83.3\%}      & \textbf{0.037/0.034}           \\ 
\cmidrule(lr){1-13}
BDETT                  & 70.7\%               & 0.104/0.109                   & 64.0\%               & 0.073/0.097                   & 79.6\%               & 0.016/0.022                   & 71.6\%               & 0.021/\textbf{0.022}          & 87.5\%               & 0.025/\textbf{0.026}          & 73.4\%               & \textbf{0.019}/0.023           \\
BDETT-DWAM             & \textbf{82.6\%}      & \textbf{0.103/0.103}          & \textbf{75.0\%}      & \textbf{0.070/0.096}          & \textbf{82.8\%}      & \textbf{0.015/0.021}          & \textbf{80.0\%}      & \textbf{0.017/0.022}          & \textbf{89.3\%}      & \textbf{0.024/0.026}          & \textbf{76.6\%}      & \textbf{0.019/0.022}           \\ 
\midrule
                       & \multicolumn{4}{c}{\textbf{8-bit~Loihi~weight }}                                                            & \multicolumn{4}{c}{\textbf{GN~weight }}                                                                     & \multicolumn{4}{c}{\textbf{30\%~Zero~weight }}                                                                \\ 
\cmidrule(lr){2-5}\cmidrule(lr){6-9}\cmidrule(lr){10-13}
                       & \multicolumn{2}{c}{LIF}                              & \multicolumn{2}{c}{SRM}                              & \multicolumn{2}{c}{LIF}                              & \multicolumn{2}{c}{SRM}                              & \multicolumn{2}{c}{LIF}                              & \multicolumn{2}{c}{SRM}                               \\ 
\cmidrule(lr){2-3}\cmidrule(lr){4-5}\cmidrule(lr){6-7}\cmidrule(lr){8-9}\cmidrule(lr){10-11}\cmidrule(lr){12-13}
\textbf{\textbf{Name}} & SR~                  & $\text{HM}_m$/$\text{HM}_{std}$ & SR~                  & $\text{HM}_m$/$\text{HM}_{std}$ & SR~                  & $\text{HM}_m$/$\text{HM}_{std}$ & SR~                  & $\text{HM}_m$/$\text{HM}_{std}$ & SR~                  & $\text{HM}_m$/$\text{HM}_{std}$ & SR~                  & $\text{HM}_m$/$\text{HM}_{std}$  \\ 
\midrule
SAN                 & 85.4\%               & 0.005/0.005                   & \textbf{85.5\%}      & \textbf{0.004/0.003}          & 55.2\%               & 0.144/0.168                   & 0.0\%                & 0.166/0.158                   & 66.3\%               & 0.102/0.104                   & 5.7\%                & 0.108/0.105                    \\
SAN-DWAM            & \textbf{90.1\%}      & \textbf{0.004/0.003}          & 84.7\%               & \textbf{0.004}/0.004          & \textbf{70.4\%}      & \textbf{0.106/0.127}          & \textbf{0.5\%}       & \textbf{0.162/0.148}          & \textbf{77.4\%}      & \textbf{0.093/0.102}          & \textbf{14.0\%}      & \textbf{0.105/0.100}           \\ 
\cmidrule(lr){1-13}
BDETT                  & 91.1\%               & \textbf{0.003/0.003}          & 83.3\%               & 0.003/\textbf{0.002}          & 79.6\%               & 0.062/0.081                   & 79.0\%               & 0.043/0.051                   & 56.7\%               & 0.057/0.070                   & 56.0\%               & 0.056/0.066                    \\
BDETT-DWAM             & \textbf{93.8\%}      & \textbf{0.003/0.003}          & \textbf{86.9\%}      & \textbf{0.002/0.002}          & \textbf{89.8\%}      & \textbf{0.052/0.073}          & \textbf{84.4\%}      & \textbf{0.035/0.042}          & \textbf{78.3\%}      & \textbf{0.046/0.057}          & \textbf{81.6\%}      & \textbf{0.054/0.059}           \\
\bottomrule
\end{tabular}
}
\end{small}
\end{table*}

We validated the effectiveness of DWAM in two types of tasks: a robot obstacle avoidance task and continuous control tasks. To further evaluate its generalization across different neuron models, we conducted experiments using both LIF and SRM neurons.

Robotic Obstacle Avoidance Task: In this task, the robot navigates from a random start to a target point in an obstacle-filled environment. A trial is successful if the robot reaches the target without collisions. Each round includes 200 trials, with the success rate (SR) defined as:
$\text{SR} = \frac{N_{\text{success}}}{N_{\text{total}}}$,
where $N_{\text{success}}$ is the number of successful trials and $N_{\text{total}} = 200$. The task is repeated over five rounds, and the final evaluation metric is the average success rate across all rounds.

Continuous Control Tasks: We used two standard OpenAI Gym environments~\cite{brockman2016openai}: HalfCheetah-v3 and Ant-v3. The goal is to control a half-mechanical cheetah or a quadruped robot to move forward over 1000 steps while maximizing forward velocity. Each round consists of 10 trials, and the highest cumulative reward is recorded per round. The task is repeated for 10 rounds, and the average score across rounds is used for evaluation.

In addition to performance metrics (SR and reward), we evaluated the homeostasis of host SNNs using two firing-rate-based measures: $\text{HM}_{m}$ and $\text{HM}_{std}$. These measures quantify the mean and variance of absolute changes in neuron firing rates across $P$ degraded-condition trials, relative to the standard condition, reflecting the stability of network activity. Details on these measures can be found in Supplementary Note 4.

\textbf{Experimental Setup} Robotic Obstacle Avoidance Task:
We selected two types of host SNNs: one without a homeostasis mechanism, the Spiking Action Network (SAN)~\cite{tang2020reinforcement}, and another with the best current homeostasis mechanism, the Bioinspired Dynamic Energy-Temporal Threshold (BDETT)~\cite{ding2022biologically}. SAN and BDETT were first trained using the spiking deep deterministic policy gradient (SDDPG) framework~\cite{tang2020reinforcement}. After training, we directly integrated DWAM into the trained models without any further retraining, resulting in SAN-DWAM and BDETT-DWAM. We then conducted comparisons within each model group (i.e., SAN vs. SAN-DWAM, and BDETT vs. BDETT-DWAM) to evaluate the effectiveness of DWAM. See Supplementary Note 5 for further training details.

Continuous Control Task: 
we selected two types of host SNNs: one without a homeostasis mechanism, the population-coded SAN (PopSAN)~\cite{tang2020deep}, and the other with the best current homeostasis mechanism, BDETT~\cite{ding2022biologically}. PopSAN is an improved version of SAN, featuring specifically designed encoders and decoders to handle high-dimensional control tasks. We construct PopSAN-DWAM and BDETT-DWAM using the same method as the obstacle avoidance task. Both PopSAN and BDETT were trained using the twin-delayed deep deterministic policy gradient (TD3) algorithm~\cite{fujimoto2018addressing}. See Supplementary Note 6 for further training details.

To verify the necessity of our modifications to the original biological formulation, we further conducted controlled comparisons by integrating BioDWAM into SAN, PopSAN, and BDETT, resulting in SAN-BioDWAM, PopSAN-BioDWAM, and BDETT-BioDWAM. These variants were included in the experiments as supplementary baselines. Their results and implementation details are provided in Supplementary Notes 5 and 6.

\subsection{Robotic Obstacle Avoidance with DWAM}
In this task, a mobile robot navigates from a random start to a random target within a given scene, using a Robot Peak light detection and ranging (RPLIDAR) sensor for obstacle detection. To evaluate DWAM’s homeostatic capability, we tested it under several degraded conditions based on a static environment: (1) dynamic obstacles, by adding moving objects; (2) degraded input, by disturbing LIDAR data using “0.2”, “6.0” and “GN” methods; and (3) weight uncertainty, by perturbing trained weights via “8-bit Loihi”, “GN weight” and “30\% zero weight.” Detailed explanations of these disturbance methods can be found in Supplementary Note 5.

\textbf{Success Rate} 
Table~\ref{tab:OA-Degraded} presents the success rate comparison between the host SNN and the host SNN integrated with DWAM under degraded conditions, evaluated on both LIF and SRM neuron models. The integration of DWAM consistently improves the success rates of the host SNN to varying degrees, except for the SAN model using SRM neurons under the 8-bit Loihi weight condition. Notably, SAN-DWAM demonstrates clear performance gains over SAN, particularly under severe input perturbations such as GN noise with LIF neurons. Similarly, BDETT-DWAM shows a consistent advantage over BDETT, especially under conditions involving significant weight degradation. These results show that DWAM effectively introduces a stability mechanism into SNNs lacking intrinsic homeostasis, enabling them to overcome challenging conditions. The observed improvements also suggest that multiple stability mechanisms can coexist within SNNs, supporting the viability of our approach.

\textbf{Homeostasis Metrics}
In biological systems, each neuron has a unique firing rate set point. When a network is in homeostasis, the firing rates of individual neurons under different conditions should remain close to their respective set points. In other words, as the network transitions from standard conditions to degraded conditions, an SNN with stronger homeostasis is expected to exhibit smaller changes in firing rates, reflected by lower values of the $\text{HM}_{m}$ and $\text{HM}_{std}$ metrics.

Table~\ref{tab:OA-Degraded} compares firing rate changes in the host SNN with and without DWAM across various degraded conditions. DWAM consistently reduces both metrics, except in the SAN-SRM model under 8-bit Loihi weights. These results support DWAM’s effectiveness in embedding biologically inspired homeostatic regulation into SNNs.

\subsection{Continuous Control with DWAM}
The continuous control tasks include HalfCheetah-v3 and Ant-v3, which vary in complexity: HalfCheetah has a 17D observation and 6D action space, while Ant uses 110D and 8D, respectively. This diversity enables a comprehensive evaluation of DWAM in continuous control. In both tasks, the objective is to control a six-joint half-mechanical cheetah (HalfCheetah) or a quadruped robot (Ant) to move forward for 1000 steps while maximizing velocity. As in the obstacle avoidance task, we tested under degraded conditions based on the baseline environment, including degraded input (disturbing observations via “MIN”, “MAX”, and “GN”) and weight uncertainty (same as in obstacle avoidance). Detailed explanations of these disturbance methods can be found in Supplementary Note 6.

\begin{table*}[t]
\centering
\caption{Quantitative performance of the Half Cheetah-v3 task under degraded conditions, where higher reward indicates better performance, and lower $\text{HM}_{m}$ and $\text{HM}_{std}$ values reflect improved homeostasis.} 
\label{tab:HC-Degraded}
\begin{small}
\resizebox{17.5cm}{!}{
\setlength\tabcolsep{1pt}
\begin{tabular}{ccccccccccccc} 
\toprule
                       & \multicolumn{4}{c}{\textbf{MIN }}                                                                               & \multicolumn{4}{c}{\textbf{MAX}}                                                                                & \multicolumn{4}{c}{\textbf{GN}}                                                                                  \\ 
\cmidrule(lr){2-5}\cmidrule(lr){6-9}\cmidrule(lr){10-13}
                       & \multicolumn{2}{c}{LIF}                                & \multicolumn{2}{c}{SRM}                                & \multicolumn{2}{c}{LIF}                                & \multicolumn{2}{c}{SRM}                                & \multicolumn{2}{c}{LIF}                                & \multicolumn{2}{c}{SRM}                                 \\ 
\cmidrule(lr){2-3}\cmidrule(lr){4-5}\cmidrule(lr){6-7}\cmidrule(lr){8-9}\cmidrule(lr){10-11}\cmidrule(lr){12-13}
\textbf{\textbf{Name}} & Reward               & $\text{HM}_m$/$\text{HM}_{std}$ & Reward               & $\text{HM}_m$/$\text{HM}_{std}$ & Reward               & $\text{HM}_m$/$\text{HM}_{std}$ & Reward               & $\text{HM}_m$/$\text{HM}_{std}$ & Reward               & $\text{HM}_m$/$\text{HM}_{std}$ & Reward               & $\text{HM}_m$/$\text{HM}_{std}$  \\ 
\midrule
PopSAN                 & 9993                 & 0.035/0.036                   & 10354                & 0.029/0.027                   & 10251                & 0.028/0.027                   & 10511                & 0.030/0.027                   & 4050                 & 0.085/0.073                   & 3641                 & 0.095/0.084                    \\
PopSAN-DWAM            & \textbf{10155}       & \textbf{\textbf{0.032/0.034}} & \textbf{10413}       & \textbf{0.026\textbf{/}0.025} & \textbf{10306}       & \textbf{0.027\textbf{/}0.026} & \textbf{10639}       & \textbf{0.027\textbf{/}0.025} & \textbf{4139}        & \textbf{0.083\textbf{/}0.072} & \textbf{3705}        & \textbf{0.093\textbf{/}0.083}  \\ 
\cmidrule(lr){1-13}
BDETT                  & 9906                 & 0.020/\textbf{0.027}          & 11050                & 0.017/0.018                   & 9756                 & 0.023/0.028                   & 10723                & 0.018/0.018                   & 3425                 & \textbf{0.057}/0.066          & 3701                 & 0.086/0.071                    \\
BDETT-DWAM             & \textbf{10003}       & \textbf{0.019/0.027}          & \textbf{11149}       & \textbf{0.016/0.017}          & \textbf{9864}        & \textbf{0.021/0.027}          & \textbf{10868}       & \textbf{0.017/0.017}          & \textbf{3498}        & \textbf{0.057/0.065}          & \textbf{4037}        & \textbf{0.076/0.061}           \\ 
\midrule
                       & \multicolumn{4}{c}{\textbf{8-bit~Loihi~weight }}                                                            & \multicolumn{4}{c}{\textbf{GN~weight }}                                                                     & \multicolumn{4}{c}{\textbf{30\%~Zero~weight }}                                                                \\ 
\cmidrule(lr){2-5}\cmidrule(lr){6-9}\cmidrule(lr){10-13}
                       & \multicolumn{2}{c}{LIF}                              & \multicolumn{2}{c}{SRM}                              & \multicolumn{2}{c}{LIF}                              & \multicolumn{2}{c}{SRM}                              & \multicolumn{2}{c}{LIF}                              & \multicolumn{2}{c}{SRM}                               \\ 
\cmidrule(lr){2-3}\cmidrule(lr){4-5}\cmidrule(lr){6-7}\cmidrule(lr){8-9}\cmidrule(lr){10-11}\cmidrule(lr){12-13}
\textbf{\textbf{Name}} & Reward~              & $\text{HM}_m$/$\text{HM}_{std}$ & Reward~              & $\text{HM}_m$/$\text{HM}_{std}$ & Reward~              & $\text{HM}_m$/$\text{HM}_{std}$ & Reward~              & $\text{HM}_m$/$\text{HM}_{std}$ & Reward~              & $\text{HM}_m$/$\text{HM}_{std}$ & Reward~              & $\text{HM}_m$/$\text{HM}_{std}$  \\ 
\midrule
PopSAN                 & 10576                & \textbf{0.003/0.003}          & 10793                & \textbf{0.003/}0.003          & 7158                 & \textbf{0.055/}0.046          & 9671                 & 0.023/0.021                   & 4467                 & 0.105/0.088                   & 7255                 & 0.066/0.055                    \\
PopSAN-DWAM            & \textbf{10651}       & \textbf{0.003/0.003}          & \textbf{10807}       & \textbf{0.003/0.002}          & \textbf{7194}                 & \textbf{0.055/0.045}          & \textbf{9882}        & \textbf{0.021/0.019}          & \textbf{4845}        & \textbf{0.103/0.087}          & \textbf{7522}        & \textbf{0.063/0.053}           \\ 
\cmidrule(lr){1-13}
BDETT                  & 10253                & \textbf{\textbf{0.002/0.002}} & 11278                & 0.005/\textbf{\textbf{0.006}} & 8885                 & 0.017/\textbf{0.017}          & 10691                & 0.015/0.013                   & 1584                 & 0.091/0.097                   & 8167                 & 0.049/\textbf{0.038}           \\
BDETT-DWAM             & \textbf{10327}       & \textbf{0.002/0.002}          & \textbf{11414}       & \textbf{0.004/0.006}          & \textbf{8962}        & \textbf{0.016/0.017}          & \textbf{11048}       & \textbf{0.014/0.010}          & \textbf{1759}        & \textbf{0.088/0.094}          & \textbf{8219}        & \textbf{0.047/0.038}           \\
\bottomrule
\end{tabular}
}
\end{small}
\end{table*}

\begin{table*}[t]
\centering
\caption{Quantitative performance of the Ant-v3 task under degraded conditions, where higher reward indicates better performance, and lower $\text{HM}_{m}$ and $\text{HM}_{std}$ values reflect improved homeostasis.} 
\label{tab:Ant-Degraded}
\begin{small}
\resizebox{17.5cm}{!}{
\setlength\tabcolsep{1pt}
\begin{tabular}{ccccccccccccc} 
\toprule
                       & \multicolumn{4}{c}{\textbf{MIN }}                                                                               & \multicolumn{4}{c}{\textbf{MAX}}                                                                                & \multicolumn{4}{c}{\textbf{GN}}                                                                                  \\ 
\cmidrule(lr){2-5}\cmidrule(lr){6-9}\cmidrule(lr){10-13}
                       & \multicolumn{2}{c}{LIF}                                & \multicolumn{2}{c}{SRM}                                & \multicolumn{2}{c}{LIF}                                & \multicolumn{2}{c}{SRM}                                & \multicolumn{2}{c}{LIF}                                & \multicolumn{2}{c}{SRM}                                 \\ 
\cmidrule(lr){2-3}\cmidrule(lr){4-5}\cmidrule(lr){6-7}\cmidrule(lr){8-9}\cmidrule(lr){10-11}\cmidrule(lr){12-13}
\textbf{\textbf{Name}} & Reward               & $\text{HM}_m$/$\text{HM}_{std}$ & Reward               & $\text{HM}_m$/$\text{HM}_{std}$ & Reward               & $\text{HM}_m$/$\text{HM}_{std}$ & Reward               & $\text{HM}_m$/$\text{HM}_{std}$ & Reward               & $\text{HM}_m$/$\text{HM}_{std}$ & Reward               & $\text{HM}_m$/$\text{HM}_{std}$  \\ 
\midrule
PopSAN                 & 3521          & 0.092/0.066                   & 5687          & 0.057/0.060                   & 4795          & \textbf{0.079}/0.077          & 5430          & 0.051/0.049                   & 2160          & 0.125/0.091                   & 2109          & 0.093/\textbf{0.080}           \\
PopSAN-DWAM            & \textbf{3794} & \textbf{0.066\textbf{/}0.060} & \textbf{5709} & \textbf{0.055\textbf{/}0.059} & \textbf{4900} & \textbf{0.079\textbf{/}0.76}  & \textbf{5581} & \textbf{0.048\textbf{/}0.047} & \textbf{2541} & \textbf{0.105\textbf{/}0.087} & \textbf{2153} & \textbf{0.092\textbf{/}0.080}  \\ 
\cmidrule(lr){1-13}
BDETT                  & 4186          & \textbf{0.039/0.061}          & 4275          & 0.058/0.052                   & 3638          & \textbf{0.049/0.071}          & 4510          & 0.064/0.063                   & 2182          & \textbf{0.070}/0.093          & 2275          & 0.090/0.073                    \\
BDETT-DWAM             & \textbf{4214} & \textbf{0.039/0.061}          & \textbf{4388} & \textbf{0.056/0.050}          & \textbf{3691} & \textbf{0.049/0.071}          & \textbf{4516} & \textbf{0.063/0.062}          & \textbf{2230} & \textbf{0.069/0.092}          & \textbf{2517} & \textbf{0.080/0.062}           \\ 
\midrule
                       & \multicolumn{4}{c}{\textbf{8-bit~Loihi~weight }}                                                            & \multicolumn{4}{c}{\textbf{GN~weight }}                                                                     & \multicolumn{4}{c}{\textbf{30\%~Zero~weight }}                                                                \\ 
\cmidrule(lr){2-5}\cmidrule(lr){6-9}\cmidrule(lr){10-13}
                       & \multicolumn{2}{c}{LIF}                              & \multicolumn{2}{c}{SRM}                              & \multicolumn{2}{c}{LIF}                              & \multicolumn{2}{c}{SRM}                              & \multicolumn{2}{c}{LIF}                              & \multicolumn{2}{c}{SRM}                               \\ 
\cmidrule(lr){2-3}\cmidrule(lr){4-5}\cmidrule(lr){6-7}\cmidrule(lr){8-9}\cmidrule(lr){10-11}\cmidrule(lr){12-13}
\textbf{\textbf{Name}} & Reward~              & $\text{HM}_m$/$\text{HM}_{std}$ & Reward~              & $\text{HM}_m$/$\text{HM}_{std}$ & Reward~              & $\text{HM}_m$/$\text{HM}_{std}$ & Reward~              & $\text{HM}_m$/$\text{HM}_{std}$ & Reward~              & $\text{HM}_m$/$\text{HM}_{std}$ & Reward~              & $\text{HM}_m$/$\text{HM}_{std}$  \\ 
\midrule
PopSAN                 & 5904          & 0.008/0.011                   & 6044          & 0.007/0.007                   & 1324          & 0.185/0.113                   & 5580          & 0.025/0.021                   & 1580          & 0.124/0.108                   & 4091          & 0.078/0.063                    \\
PopSAN-DWAM            & \textbf{5948} & \textbf{0.007/0.010}          & \textbf{6050} & \textbf{0.006/0.005}          & \textbf{1519} & \textbf{0.145/0.097}          & \textbf{5627} & \textbf{0.021/0.019}          & \textbf{1873} & \textbf{0.122/0.102}          & \textbf{4219} & \textbf{0.077/0.062}           \\ 
\cmidrule(lr){1-13}
BDETT                  & 4732          & \textbf{0.003}/0.004          & 5732          & 0.008/0.010                   & 3626          & 0.023/\textbf{0.028}          & 4939          & 0.040/0.035                   & 1937          & 0.055/0.072                   & 4949          & 0.040/0.035                    \\
BDETT-DWAM             & \textbf{4774} & \textbf{0.003/0.003}          & \textbf{5822} & \textbf{0.007/0.009}          & \textbf{3637} & \textbf{0.022/0.028}          & \textbf{5558} & \textbf{0.017/0.015}          & \textbf{2157} & \textbf{0.053/0.071}          & \textbf{5030} & \textbf{0.037/0.033}           \\
\bottomrule
\end{tabular}
}
\end{small}
\end{table*}

\textbf{Reward} 
The reward comparison between the host SNN and the host SNN integrated with DWAM under degraded conditions for the HalfCheetah-v3 and Ant-v3 tasks is shown in Table~\ref{tab:HC-Degraded} and~\ref{tab:Ant-Degraded}. The comparison was conducted using both the LIF and SRM neuron models. Across all degraded conditions, the cumulative rewards of the host SNN show clear improvements after integrating DWAM. Specifically, SAN-DWAM consistently outperforms SAN, with the most noticeable gains observed under challenging input perturbations based on the LIF model. Similarly, BDETT-DWAM achieves higher rewards than BDETT, particularly under severe weight degradation using the SRM model. These findings, combined with the earlier results from the obstacle avoidance task, further validate the generalization capability of DWAM. It not only enhances performance in specific tasks but also demonstrates robustness across diverse task domains.

\textbf{Homeostasis Metrics} 
The impact of various degraded conditions on the firing rate changes of the host SNN and the host SNN integrated with DWAM in the HalfCheetah-v3 and Ant-v3 tasks are shown in Table~\ref{tab:HC-Degraded} and~\ref{tab:Ant-Degraded}, respectively. In all degraded conditions, the homeostasis metrics of the host SNN showed consistent reductions following the incorporation of DWAM. 

These results further confirm that the proposed DWAM is an effective method for introducing biologically-inspired homeostatic mechanisms into SNNs. By minimizing deviations in neuronal firing rates, it stabilizes network performance, even under challenging conditions.

\section{Conclusion}
In this paper, we proposed the DWAM as a biologically-inspired approach to integrating homeostatic mechanisms into SNNs. Drawing on the principles of the BCM theory, DWAM enables SNNs to dynamically regulate synaptic strength, maintaining network stability across varying levels of neuronal activity. We evaluated the proposed method across multiple tasks and under various degradation conditions. The experimental results demonstrated that DWAM not only enhances the performance of SNNs lacking homeostatic mechanisms in challenging scenarios but also seamlessly integrates with existing homeostatic methods, improving the performance of host SNNs already equipped with such features. This method broadens the scope of SNN design, offering a novel perspective for enhancing network robustness and adaptability in a biologically plausible manner. Nevertheless, the interplay and joint effects of multiple homeostatic mechanisms within SNNs remain poorly understood. If such mechanisms could not only coexist but also complement each other as they do in biological systems, more effective and resilient forms of homeostasis might be achieved — a promising direction for future research.

\bibliography{aaai2026}

\clearpage

\input{supplementary.tex}

\end{document}

%% file: supplementary.tex
\onecolumn
\section*{Supplementary Note 1: Bioinspired Dynamic Energy-Temporal
Threshold (BDETT)}
The BDETT provides a dynamic adjustment of neuron spiking thresholds in Spiking Neural Networks (SNNs) to emulate biological properties. This method draws inspiration from two bioplausible observations:

\textbf{Positive Correlation with Average Membrane Potential:} The threshold increases as the neuron’s average membrane potential grows, stabilizing the spiking activity in the presence of sustained input.

\textbf{Negative Correlation with Preceding Depolarization Rate:} The threshold decreases when the membrane potential rises rapidly, enhancing the neuron’s sensitivity to sudden input changes.

To mathematically define the dynamic spiking threshold, the BDETT uses the following formulation~\cite{ding2022biologically}:
\begin{equation}
\Theta_i^{l}(t+1) = \frac{1}{2}({\rm E}_i^l(t) + {\rm T}_i^l(t+1)),
\end{equation}
where $E_i^l(t)$ is the dynamic energy threshold (DET) of the neuron for ensuring a positive correlation, and $T_i^l(t+1)$ is the dynamic temporal threshold (DTT), which ensures a negative correlation.

DET: For the $i$-th neuron in the $l$-th layer at timestamp $t$, BDETT defines DET as~\cite{ding2022biologically}:
\vspace{-0.2 cm}
\begin{align}
{\rm E}^{l}_i(t) &= \eta (v^{l}_i(t) - V_m^{l}(t)) + V_\theta^{l}(t) + ln(1 + e^{\frac{v^{l}_i(t) - V_m^{l}(t)}{\psi}}), \label{eq:E}\\
V_m^{l}(t) &= \mu(v_i^l(t)) - 0.2(\max(v_i^{l}(t)) - \min(v_i^{l}(t)))\quad \text{for } i= 1,2, ..., n^l,\label{eq:E_vm}\\
V_\theta^{l}(t) &= \mu(\Theta_i^{l}(t)) - 0.2(\max(\Theta_i^{l}(t)) - \min(\Theta_i^{l}(t))) \quad \text{for } i= 1,2, ..., n^l,\label{eq:E_vt}
\end{align}
where $v^{l}_i(t)$ is the neuron postsynaptic membrane potential at timestamp $t$; $\mu$ is the mean operator; $n^l$ is the total number of neurons in the $l$-th layer; and $\eta$ and $\psi$ are two hyperparameters, which are set empirically.

DTT: For the $i$-th neuron in the $l$-th layer, the DTT at timestamp $t+1$ is defined as~\cite{ding2022biologically}:
\vspace{-0.1cm}
\begin{align}
{\rm T}_i^{l}(t+1) &= a + e^{\frac{-(v_i^{l}(t+1) - v_i^{l}(t))}{C}}, \label{eq:G}\\
a &= -e^{-|\mu(\Theta_i^l(t))|}\quad \text{for } i=1,2,...,n^l.
\end{align}

This dynamic threshold design ensures that neurons achieve homeostasis by adapting their firing thresholds according to input patterns. 

\section*{Supplementary Note 2: Spiking Neural Networks}
\subsection*{Spiking Neural Networks (SNNs)}

SNNs are a class of biologically inspired neural networks that aim to replicate the information processing mechanisms of biological neurons. Unlike traditional artificial neural networks, SNNs communicate through discrete spike events rather than continuous signals. These spikes are generated when the membrane potential of a neuron crosses a certain threshold, resulting in an action potential, a process known as firing. The state of a neuron in SNNs is typically categorized into resting, depolarization, and hyperpolarization. In the resting state, the neuron's membrane potential remains stable at a baseline level. Depolarization refers to an increase in membrane potential, making the neuron more likely to fire, while hyperpolarization describes a decrease, reducing the likelihood of firing. 

A sequence of spikes is called a spike train and is defined as $s(t)=\Sigma_{t^{(f)}\in \mathcal{F}}\delta(t-t^{(f)})$, where $\mathcal{F}$ represents the set of times at which the individual spikes occur~\cite{shrestha2018slayer}. When an input spike train reaches a neuron, it influences the membrane potential through a combination of integration and decay processes. If the membrane potential exceeds a predefined threshold, the neuron emits an output spike, which in turn propagates to downstream neurons. This mechanism is the foundation of SNN computation. Over the years, various neuron models have been developed to simulate the dynamics of spiking neurons, with two commonly used ones being the Spike Response Model (SRM)~\cite{gerstner1995time} and the Leaky Integrate-and-Fire (LIF) model~\cite{gerstner2002spiking}. 

\textbf{Spike Response Model (SRM)}
The SRM describes a neuron's membrane potential as a dynamic response to incoming spikes. When an input spike train $s_i(t)$ arrives at a neuron, each spike triggers a response characterized by a kernel function $\epsilon(t)$, representing the temporal evolution of the depolarization process. The cumulative depolarization of the neuron is given by: $\Sigma_i w_i(\varepsilon \ast s_i)(t)$, where $w_i$ is the synaptic weight of the $i$-th input.  

The SRM also incorporates the process of hyperpolarization by introducing a refractory kernel $\zeta(t)$. This kernel accounts for the temporary suppression of the neuron's ability to fire after a spike, effectively modeling the hyperpolarization phase: $(\zeta \ast s)(t)$, where the second term reduces the membrane potential after a spike, preventing immediate firing.  

With an SRM, a feedforward SNN architecture with $n_l$ layers can be defined. Given $N^l$ incoming spike trains at layer $l$, $s_i^l(t)$, the forward propagation process of the network is mathematically defined as follows~\cite{gerstner1995time, shrestha2018slayer}: 

\begin{align}
v_i^{l+1}(t) &= \sum_{j=1}^{N^{l}}w_{ij}(\varepsilon \ast s_j^{l})(t) + (\zeta \ast s_i^{l+1})(t-1) ,  \\
s_i^{l+1}(t) &= f_s(v_i^{l+1}(t)), \label{eq:srm_s}\\
f_s(v) &: v \rightarrow s, s(t):= s(t) + \delta(t-t^{(f+1)}), \label{eq:srm_fs}\\
t^{f+1} &= \min\{t: v(t) = \Theta, t > t^{(f)}\}, \label{eq:srm_TH}
\end{align}
where $f_s(\cdot)$ is a spike function and $\Theta$ is the membrane potential threshold, which is static and the same for all neurons in the network. This static threshold is the one that we replace with the proposed bioplausible dynamic threshold. 

\textbf{Leaky Integrate-and-Fire (LIF)}
The LIF model is a simplified variant of the SRM. Its depolarization process is similar to that of the SRM, but it directly processes incoming spike trains while ignoring the spike response kernel. After the neuron fires, the membrane potential is immediately reset to a lower value, typically modeled as 0 or the resting potential, marking the onset of the hyperpolarization phase.

The forward propagation process of the network can be defined as:

\begin{align}
v_{i}^{l+1}(t) &=  \sum_{j=1}^{N^{l}}w_{ij}s_j^{l}(t) + v_{i}^{l+1}(t-1)f_d(s_{i}^{l+1}(t-1)) + b_i^{l+1}, \label{eq:lif_v}\\
s_i^{l+1}(t) &= f_s(v_i^{l+1}(t)), \label{eq:lif_s} \\
f_d(s(t)) &= \begin{cases}
                  D& \quad s(t) = 0\\
                  0& \quad s(t) = 1,
\end{cases}
\end{align}
where $b_i^{l+1}$ is an adjustable bias that is learned to mimic a dynamic threshold behavior. However, the biases of this model are static during forwarding propagation. In contrast, the proposed dynamic threshold is dynamic and automatically adapts to membrane potentials. 

\section*{Supplementary Note 3: Details of the toy example}
The network structure and training process used in the toy example are identical to those in the robotic obstacle avoidance experiments (details can be found in Supplementary Note 5). However, DWAM is applied only to the neurons controlling linear velocity and their associated weights, with the firing rate and modification threshold tracked over time. The experimental scenario is a simplified version of the obstacle avoidance task, where all static and dynamic obstacles have been removed. In the absence of obstacles, the firing frequency of the linear velocity neurons approaches 1 to provide higher linear velocity, allowing the robot to reach the target point more quickly.

\section*{Supplementary Note 4: Homeostasis Metrics}
We optimized the calculation of the homeostasis metrics. Existing work defines the homeostasis metrics as~\cite{ding2022biologically}: 
\begin{equation}
\begin{aligned}
\text{FR}_m &= \mu(\text{FR}_m^p)\quad \text{for}\quad p=1,2,..., P,\\
\text{FR}_{std}^m &= \mu(\text{FR}_{std}^p)\quad \text{for}\quad p=1,2,...,P,\\
\text{FR}_{std}^s &= \sigma(\text{FR}_{std}^p)\quad \text{for}\quad p=1,2,...,P,\\
\text{FR}_m^p &= \mu(f_i^{l,p}) \quad \text{for } i= 1,2, ..., N^l\quad l= 1,2, ..., L, \\
\text{FR}_{std}^p &= \sigma(f_i^{l,p}) \quad \text{for } i= 1,2, ..., N^l\quad l= 1,2, ..., L, \\
f^{l,p}_i &= \frac{\sum_{t^p=1}^{T^p}s_i^l(t^p)}{T^p} \label{eq:ding_homeostasis}
\end{aligned}
\end{equation}
where, $T^p$ is the time taken for the $p$-th trial and $f_i^{l,p}$ is the firing rate of the $i$-th neuron in the $l$-th layer during the $p$-th trial; $\text{FR}_m^p$ denotes the mean firing rate of all neurons of an SNN during the $p$-th trial, and $\text{FR}_{std}^p$ is the standard deviation of all neuron firing rates for an SNN during the $p$-th trial; $\text{FR}_m$ is the mean neuron firing rate of an SNN across all $P$ trials; $\text{FR}_{std}^m$ is the average of $P$ standard deviations, and each of them is the standard deviation of the neuron firing rates of an SNN during a single trial; $\text{FR}_{std}^s$ denotes the standard deviation of the $P$ standard deviations; $\text{FR}_{std}^s$ represents the standard deviation across all $P$ trials, while $\text{FR}_{std}^m$ denotes the mean of these standard deviations.
And the homeostasis of the network is measured by the difference between the homeostasis metrics in the base condition and the homeostasis metrics in the degraded condition. This method measures the homeostasis of the network as a whole by taking the mean (standard deviation) and then the difference, but it ignores the homeostasis of individual neurons. 

Consider a simple case: two trials were conducted on a network with four neurons $n_{1}$, $n_{2}$, $n_{3}$, $n_{4}$ under both base and degraded conditions. In the base condition, the firing rates of the four neurons in the two trials are $f^{p=1}_{i=1, 2, 3, 4}=(0.3, 0.5, 0.5, 0.7)$ and $f^{p=2}_{i=1, 2, 3, 4}=(0.2, 0.5, 0.5, 0.8)$; In the degraded condition, the firing rates of the four neurons in the two trials are $f^{p=1}_{i=1, 2, 3, 4}=(0.7, 0.5, 0.5, 0.3)$ and $f^{p=2}_{i=1, 2, 3, 4}=(0.8, 0.5, 0.5, 0.2)$. According to Eq.~\ref{eq:ding_homeostasis}, $\Delta \text{FR}_m$, $\Delta \text{FR}_{std}^m$ and $\Delta \text{FR}_{std}^s$ are equal to 0, suggesting the network has an optimal homeostasis. However, in reality, the firing rates of $n_{1}$ and $n_{4}$ fluctuate significantly.

In biology, each neuron has its unique firing rate set-point and homeostasis~\cite{giachello2017regulation}. In the biological experiment~\cite{driscoll2013pumilio}, the homeostasis is calculated by first taking the difference and then averaging (or finding the standard deviation), which can adequately measure the steady state of each neuron. Therefore, we optimized the calculation of the homeostasis metrics, defined as follows:
\begin{equation}
\begin{aligned}
\text{HM}_m &= \mu(abs(f^{l,p_{base}}_{i}-f^{l,p_{degraded}}_{i}))\quad \text{for}\quad p=1,2,..., P,\\
\text{HM}_{std} &= \sigma(abs(f^{l,p_{base}}_{i}-f^{l,p_{degraded}}_{i}))\quad \text{for}\quad p=1,2,..., P,\\
f^{l,p}_i &= \frac{\sum_{t^p=1}^{T^p}s_i^l(t^p)}{T^p} \label{eq:BDHFS_homeostasis}
\end{aligned}
\end{equation}
where $f^{l,p}_i$, $\mu$, and $\sigma$ are defined in the same way as $f^{l,p}_i$ in Eq.~\ref{eq:ding_homeostasis}; $abs()$ is taken as an absolute value; $f^{l,p_{base}}_{i}$ and $f^{l,p_{degraded}}_{i}$ are the firing rates of the $i$-th neuron in the $l$-th layer during the $p$-th trial in the base and degraded conditions, respectively.

\section*{Supplementary Note 5: Additional Details on Obstacle Avoidance Experiments}
In this task, we control a mobile robot to navigate from a random starting point to a random target point within a given scene. The robot is equipped with a Robot Peak light detection and ranging (RPLIDAR) system, which serves as the obstacle detection sensor. It provides a 180-degree field of view composed of 18 range measurements. To evaluate DWAM’s ability to provide homeostasis to the network, we conducted experiments under various degraded conditions based on the baseline static environment. These conditions include the dynamic obstacles condition, degraded input conditions, and weight uncertainty conditions. 

Dynamic Obstacles: 11 dynamic obstacles were added to the baseline static environment. These obstacles continuously move between two points. Since these obstacles were absent during training, this condition evaluates DWAM’s ability to maintain homeostasis in a constantly changing environment.

Degraded Input: Building on the dynamic obstacles condition, the robot’s LIDAR range data was perturbed using three methods: “0.2”: The range measurements of the 3rd, 9th, and 15th beams were set to 0.2 meters, reporting obstacles regardless of their actual presence. “6.0”: The range measurements of the same beams were set to 6.0 meters (the average visible range in the test environment), making them unable to detect any obstacles. “GN”: Gaussian noise~\cite{choi2019deep}, $\mathcal N(0,1.0)$ , was added to all 18 range measurements. These conditions test DWAM’s ability to maintain homeostasis in the presence of unstable input signals.

Weight Uncertainty: Building on the dynamic obstacles condition, The robot’s learned weights from the training phase were perturbed using three methods:“8-bit Loihi weight”: Weights were converted to 8-bit integers to simulate precision loss when deploying SNNs on neuromorphic chips. “GN weight”: Gaussian noise $\mathcal N(0,0.05)$ was added to each weight. “30\% zero weight”: Randomly setting 30\% of the weights to zero in each layer.These scenarios evaluate DWAM’s ability to maintain homeostasis in the presence of unstable network parameters.

We selected two types of host SNNs: one without a homeostasis mechanism, the Spiking Action Network (SAN)~\cite{tang2020reinforcement}, and another with the best current homeostasis mechanism, the Bioinspired Dynamic Energy-Temporal Threshold (BDETT)~\cite{ding2022biologically}. Due to the similarity of the tasks, we used almost the same experimental and training setup as Ding.
\subsection*{Experimental Setup}
In the obstacle avoidance experiments, our evaluation baseline model and test environment are the SAN and its original simulated test environment, respectively. The SAN is a part of the spiking deep deterministic policy gradient (SDDPG) framework~\cite{tang2020reinforcement}, which is a fully connected four-layer SNN (i.e., three 256-neuron hidden layers and one two-neuron output layer). This network maps a state $s$ of a robot to a control action $a$.
Specifically, a state $s = \{G_{dis}, G_{dir}, \nu, \omega, L\}$ is encoded into $24$ Poisson spike trains as inputs of the SAN, and each spike train has $T$ timesteps. $G_{dis}$ and $G_{dir}$ are the relative $1$-D distances from the robot to the goal and a $2$-D direction (i.e., right and left directions), respectively; $\nu$ and $\omega$ are the robot's $1$-D linear and $2$-D angular velocities (i.e., rightward and leftward angular velocities); $L$ denotes the distance measurements obtained from a Robo Peak light detection and ranging (RPLIDAR) laser range scanner (range: {0.2-40 m}), which has a field of view of $180$ degrees with $18$ range measurements, each with a $10$-degree resolution. The two output spike trains represent the robot’s linear and angular velocities, respectively, and are decoded to control the robot via an action $a = \{\nu_L, \nu_R\}$, where $\nu_L$ and $\nu_R$ are the left and right wheel speeds of the differential-drive mobile robot, respectively~\cite{tang2020reinforcement}. 

BDETT builds upon the implementation of SAN by replacing its firing threshold with the threshold calculated using the BDETT method~\cite{ding2022biologically}. The specific calculation details can be found in Supplementary Note 1. 

\begin{figure}[h]
	\centering
	\includegraphics [scale=0.5]{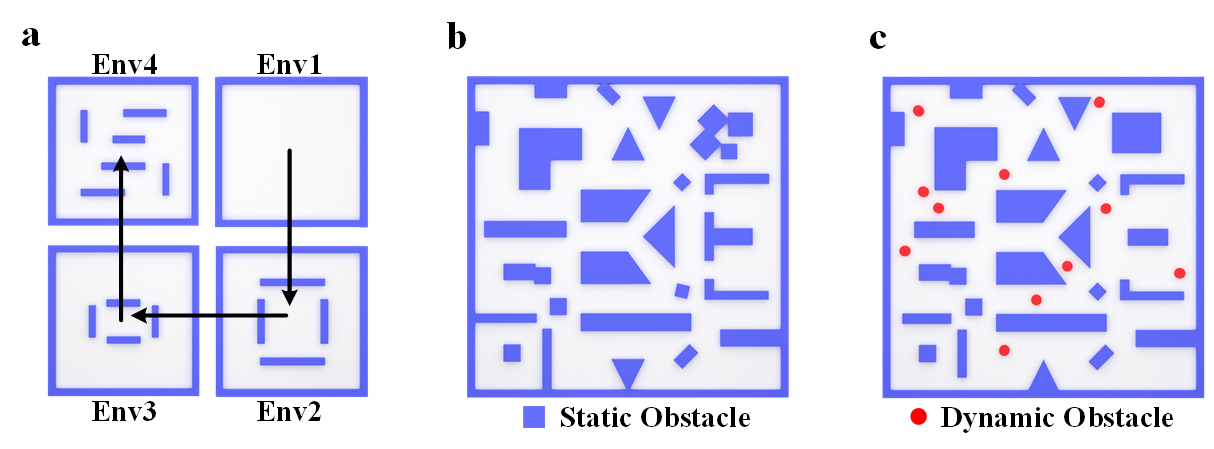}
	\vspace{-0.2cm}
	\caption{
	    Illustrations of the training, static testing and dynamic testing environments.
		a. The training environments of the obstacle avoidance tasks. The training processes of all competing SNNs start from Env1 and end with Env4. b. Static testing environment. c. Dynamic testing environment. In addition to static obstacles, $11$ dynamic obstacles are inserted. 
	}
	\label{SMfig:env}
\end{figure}

\subsection*{Training Setup} 
The SAN and BDETT are trained with the original SDDPG framework. 
The training environments consist of four different maps, as shown in Figure~\ref{SMfig:env}a. In particular, during the training process, we set $100$, $200$, $300$, and $400$ start-goal pairs in the Env1, Env2, Env3, and Env4, respectively. The training starts from Env1 and ends with Env4. Following the training protocol described by \citet{tang2020reinforcement}, the hyperparameters related to training are set as follows: $D=0.75$ for the LIF; $\eta=0.01$ and $\psi=4.0$ for the dynamic energy threshold (DET); $C=3.0$ for the dynamic temporal threshold (DTT); $\tau_s=\tau_r=1.0$ for the SRM; $\text{collision reward}=-20$; $\text{goal reward}=30$; $\text{step reward}=15$; goal $l_2$ distance $\text{threshold}=0.5$ m; obstacle $l_2$ distance $\text{threshold}=0.35$ m; $\epsilon$ ranges for Env1 to Env4 of $(0.9, 0.1)$, $(0.6, 0.1)$, $(0.6, 0.1)$, and $(0.6, 0.1)$, respectively; and corresponding $\epsilon$-decays of $0.999$ for the four environments. During the training procedure, we set the batch size to $256$ and the learning rates to $0.00001$ for both the actor and critic networks.

After training SAN and BDETT, we directly integrated BioDWAM and DWAM into the trained models without any further retraining, resulting in SAN-BioDWAM, SAN-DWAM, BDETT-BioDWAM, and BDETT-DWAM. We then conducted comparisons within each model group (i.e., SAN vs. SAN-DWAM vs. SAN-BioDWAM, and BDETT vs. BDETT-DWAM vs. BDETT-BioDWAM) to evaluate the effectiveness of DWAM. The hyperparameters of BioDWAM and DWAM were set as follows: $\psi$ = 0.00005, $\alpha$ = 0.5, $\zeta$ = 2.0, and $n$ = 5.0. 

Notably, Ding et al. only provided the source code for the BDETT method based on the LIF model, including the network architecture and testing phase. As a result, we reproduced the BDETT implementation based on the SRM model by referring to the available LIF-based code. Moreover, due to the lack of training code, we reimplemented the training procedure following the protocol described by Ding et al.~\cite{ding2022biologically}. In certain experiments, this reproduction process may have led to lower BDETT performance compared to that reported in the original paper. However, since BDETT-DWAM is built upon BDETT and our experimental setup is designed to compare the two under identical conditions, any performance degradation would affect both methods equally. Therefore, such degradation is offset and does not affect the validity of our evaluation of DWAM.

We use PyTorch~\cite{paszke2019pytorch} to train and test all competing SNNs with an i7-14700K CPU and an NVIDIA GTX 4090 GPU. We direct the readers to the SAN algorithm~\cite{tang2020reinforcement} and BDETT algorithm~\cite{ding2022biologically} for details.

\subsection*{Experimental Results}
Table~\ref{SMtab:Sta&Dyn} presents the success rates and homeostasis metrics of various methods in both static and dynamic scenarios. $\sigma$ represents the variance. Table~\ref{SMtab:OA_Degraded} presents the success rates and homeostasis metrics of various methods under different degraded conditions.

\begin{table}[ht]
\centering
\caption{Quantitative performance of obstacle avoidance in static and dynamic obstacle scenarios.}
\label{SMtab:Sta&Dyn}
\vskip 0.15in
\begin{small}
\begin{tabular}{cccc} 
\toprule
                                                                            &               & \textbf{LIF}~                  & \textbf{SRM}                    \\ 
\cmidrule(lr){3-3}\cmidrule(lr){4-4}
\textbf{Type}                                                               & \textbf{Name} & SR ($\sigma$)                  & SR ($\sigma$)                   \\ 
\midrule
\multirow{6}{*}{\begin{tabular}[c]{@{}c@{}}Static\\obstacle~\end{tabular}}  & SAN           & 97.5\% (1.0)                   & 97.3\% (1.2)                    \\
                                                                            & SAN-BioDWAM   & 91.9\% (1.7)                   & 83.3\% (3.4)                    \\
                                                                            & SAN-DWAM      & \textbf{98.4\%} (0.9)          & \textbf{98.5\%} (1.3)           \\ 
\cmidrule(lr){2-4}
                                                                            & BDETT         & 97.7\% (0.5)                   & 98.3\% (0.9)                    \\
                                                                            & BDETT-BioDWAM & 78.0\% (1.9)                   & 38.3\% (3.2)                    \\
                                                                            & BDETT-DWAM    & \textbf{98.4\%} (1.2)          & \textbf{98.4\%} (0.9)           \\ 
\midrule
\multirow{6}{*}{\begin{tabular}[c]{@{}c@{}}Dynamic \\obstacle\end{tabular}} & SAN           & 85.9\% (2.4)                   & 87.3\% (2.5)                    \\
                                                                            & SAN-BioDWAM   & 81.5\% (3.6)                   & 67.2\% (2.9)                    \\
                                                                            & SAN-DWAM      & \textbf{\textbf{90.1\%}} (1.3) & \textbf{\textbf{88.9\% }}(1.9)  \\ 
\cmidrule(lr){2-4}
                                                                            & BDETT         & 91.1\% (1.5)                   & 80.1\% (2.9)                    \\
                                                                            & BDETT-BioDWAM & 70.8\% (7.0)                   & 42.7\% (1.9)                    \\
                                                                            & BDETT-DWAM    & \textbf{\textbf{92.0\% }}(1.8) & \textbf{\textbf{87.3\% }}(1.6)  \\
\bottomrule
\end{tabular}
\end{small}
\vskip -0.1in
\end{table}

\begin{table}[ht]
\centering
\caption{Quantitative performance and homeostasis metrics of obstacle avoidance under various degraded conditions.}
\label{SMtab:OA_Degraded}
\vskip 0.15in
\begin{small}
\begin{tabular}{cccccccc} 
\toprule
                                                                             &               & \multicolumn{3}{c}{\textbf{LIF}~}                           & \multicolumn{3}{c}{\textbf{SRM}}                              \\ 
\cmidrule(l){3-8}
\textbf{Type}                                                                & \textbf{Name} & SR ($\sigma$)         & $\text{HM}_m$    & $\text{HM}_{std}$  & SR ($\sigma$)          & $\text{HM}_m$    & $\text{HM}_{std}$   \\ 
\midrule
\multirow{6}{*}{0.2~}                                                        & SAN           & 88.2\% (1.0)          & 0.10895          & 0.11510          & 86.0\% (0.6)           & 0.12981          & 0.11743           \\
                                                                             & SAN-BioDWAM   & 83.9\% (1.9)          & 0.11736          & 0.11677          & 66.4\% (3.0)           & 0.13217          & 0.11770           \\
                                                                             & SAN-DWAM      & \textbf{90.6\%} (1.0) & \textbf{0.09109} & \textbf{0.11282} & \textbf{87.3\% }(3.2)  & \textbf{0.12123} & \textbf{0.11233}  \\ 
\cmidrule(lr){2-8}
                                                                             & BDETT         & 70.7\% (2.4)          & 0.10358          & 0.10895          & 64.0\% (3.1)           & 0.07301          & 0.09742           \\
                                                                             & BDETT-BioDWAM & 58.1\% (1.2)          & 0.10476          & 0.10410          & 27.6\% (2.8)           & 0.07530          & 0.09635           \\
                                                                             & BDETT-DWAM    & \textbf{82.6\% }(3.2) & \textbf{0.10331} & \textbf{0.10331} & \textbf{75.0\%} (1.5)  & \textbf{0.07028} & \textbf{0.09562}  \\ 
\midrule
\multirow{6}{*}{6.0}                                                         & SAN           & 71.7\% (2.4)          & 0.01844          & 0.02180          & 75.6\% (2.9)           & 0.02707          & 0.02807           \\
                                                                             & SAN-BioDWAM   & 75.3\% (7.0)          & 0.01888          & 0.02219          & 57.3\% (2.5)           & 0.02719          & 0.02799           \\
                                                                             & SAN-DWAM      & \textbf{84.3\%} (2.8) & \textbf{0.01827} & \textbf{0.02162} & \textbf{76.1\%} (2.4)  & \textbf{0.02629} & \textbf{0.02740}  \\ 
\cmidrule(lr){2-8}
                                                                             & BDETT         & 79.6\% (2.0)          & 0.01593          & 0.02198          & 71.6\% (3.7)           & 0.02079          & 0.02239           \\
                                                                             & BDETT-BioDWAM & 51.2\% (2.4)          & 0.01775          & 0.02180          & 29.3\% (2.5)           & 0.02198          & 0.02234           \\
                                                                             & BDETT-DWAM    & \textbf{82.8\% }(4.7) & \textbf{0.01521} & \textbf{0.02147} & \textbf{80.0\%} (2.8)  & \textbf{0.01744} & \textbf{0.02205}  \\ 
\midrule
\multirow{6}{*}{GN}                                                          & SAN           & 87.6\% (4.1)          & 0.03307          & 0.03137          & 81.6\% (3.6)           & 0.03944          & 0.03419           \\
                                                                             & SAN-BioDWAM   & 82.8\% (1.4)          & 0.03400          & 0.03149          & 68.8\% (1.5)           & 0.03792          & 0.03420           \\
                                                                             & SAN-DWAM      & \textbf{88.4\%} (3.9) & \textbf{0.03057} & \textbf{0.03049} & \textbf{83.3\%} (2.5)  & \textbf{0.03728} & \textbf{0.03381}  \\ 
\cmidrule(lr){2-8}
                                                                             & BDETT         & 87.5\% (2.0)          & 0.02464          & 0.02598          & 73.4\% (3.0)           & 0.01941          & 0.02268           \\
                                                                             & BDETT-BioDWAM & 63.2\% (2.9)          & 0.02729          & 0.02622          & 47.1\% (8.5)           & 0.02357          & 0.02339           \\
                                                                             & BDETT-DWAM    & \textbf{89.3\%} (1.4) & \textbf{0.02374} & \textbf{0.02572} & \textbf{76.6\%} (2.2)  & \textbf{0.01917} & \textbf{0.02199}  \\ 
\midrule
\multirow{6}{*}{\begin{tabular}[c]{@{}c@{}}8-bit Loihi\\weight\end{tabular}} & SAN           & 85.4\% (2.0)          & 0.00525          & 0.00499          & \textbf{85.5\%} (3.7)           & \textbf{0.00363} & \textbf{0.00346}  \\
                                                                             & SAN-BioDWAM   & 81.3\% (1.5)          & 0.00468          & 0.00440          & 82.9\% (4.0)           & 0.03535          & 0.03891           \\
                                                                             & SAN-DWAM      & \textbf{90.1\%} (1.2) & \textbf{0.00378} & \textbf{0.00344} & 84.7\% (5.1)  & 0.00392          & 0.00362           \\ 
\cmidrule(lr){2-8}
                                                                             & BDETT         & 91.1\% (0.5)          & 0.00279          & 0.00284          & 83.3\% (1.4)           & 0.00270          & 0.00238           \\
                                                                             & BDETT-BioDWAM & 65.8\% (1.0)          & 0.00319          & 0.00277          & 34.2\% (1.7)           & 0.01475          & 0.01062           \\
                                                                             & BDETT-DWAM    & \textbf{93.8\%} (0.2) & \textbf{0.00275} & \textbf{0.00265} & \textbf{86.9\%} (1.9)  & \textbf{0.00215} & \textbf{0.00223}  \\ 
\midrule
\multirow{6}{*}{\begin{tabular}[c]{@{}c@{}}GN\\weight\end{tabular}}          & SAN           & 55.2\% (6.4)          & 0.14374          & 0.16831          & 0.0\% (0.0)            & 0.16573          & 0.15850           \\
                                                                             & SAN-BioDWAM   & 0.7\% (1.2)           & 0.15385          & 0.16528          & 0.0\% (0.0)            & 0.16647          & 0.16080           \\
                                                                             & SAN-DWAM      & \textbf{70.4\% }(0.7) & \textbf{0.10598} & \textbf{0.12725} & \textbf{0.5\% }(1.0)   & \textbf{0.16165} & \textbf{0.14818}  \\ 
\cmidrule(lr){2-8}
                                                                             & BDETT         & 79.6\% (1.6)          & 0.06210          & 0.08107          & 79.0\% (1.9)           & 0.04273          & 0.05052           \\
                                                                             & BDETT-BioDWAM & 53.5\% (4.3)          & 0.05816          & 0.07334          & 30.3\% (2.9)           & 0.04603          & 0.04489           \\
                                                                             & BDETT-DWAM    & \textbf{89.8\%} (1.7) & \textbf{0.05189} & \textbf{0.07328} & \textbf{84.4\%} (2.7)  & \textbf{0.03500} & \textbf{0.04207}  \\ 
\midrule
\multirow{6}{*}{\begin{tabular}[c]{@{}c@{}}~30\% Zero \\weight\end{tabular}}  & SAN           & 66.3\% (13.2)         & 0.10150          & 0.10449          & 5.7\% (3.9)            & 0.10763          & 0.10460           \\
                                                                             & SAN-BioDWAM   & 44.3\% (13.5)         & 0.09764          & 0.10746          & 7.5\% (5.7)            & 0.10776          & 0.10445           \\
                                                                             & SAN-DWAM      & \textbf{77.4\%} (2.0) & \textbf{0.09283} & \textbf{0.10234} & \textbf{14.0\%} (14.7) & \textbf{0.10545} & \textbf{0.10042}  \\ 
\cmidrule(lr){2-8}
                                                                             & BDETT         & 56.7\% (9.9)          & 0.05747          & 0.07007          & 56.0\% (4.8)           & 0.05638          & 0.06649           \\
                                                                             & BDETT-BioDWAM & 37.6\% (6.5)          & 0.05318          & 0.06247          & 25.7\% (3.3)           & 0.05862          & 0.06159           \\
                                                                             & BDETT-DWAM    & \textbf{78.3\%} (2.9) & \textbf{0.04580} & \textbf{0.05699} & \textbf{81.6\%} (2.5)  & \textbf{0.05351} & \textbf{0.05937}  \\
\bottomrule
\end{tabular}
\end{small}
\vskip -0.1in
\end{table}

\subsection*{Understanding the Functional Role of DWAM}

To investigate how DWAM contributes to network robustness, we randomly selected one trajectory from the obstacle avoidance task as a toy example. Both SAN and SAN-DWAM models were tested along the same start-to-goal path under noise-free and noisy(GNW) conditions. During each run, we recorded the firing rates of all hidden-layer neurons at each time step. These recordings yielded four groups of neuronal activity: (1) SAN (no noise), (2) SAN (with noise), (3) SAN-DWAM (no noise), and (4) SAN-DWAM (with noise), which were then analyzed using PCA.

\begin{figure}[h]
	\centering
	\includegraphics [scale=0.5]{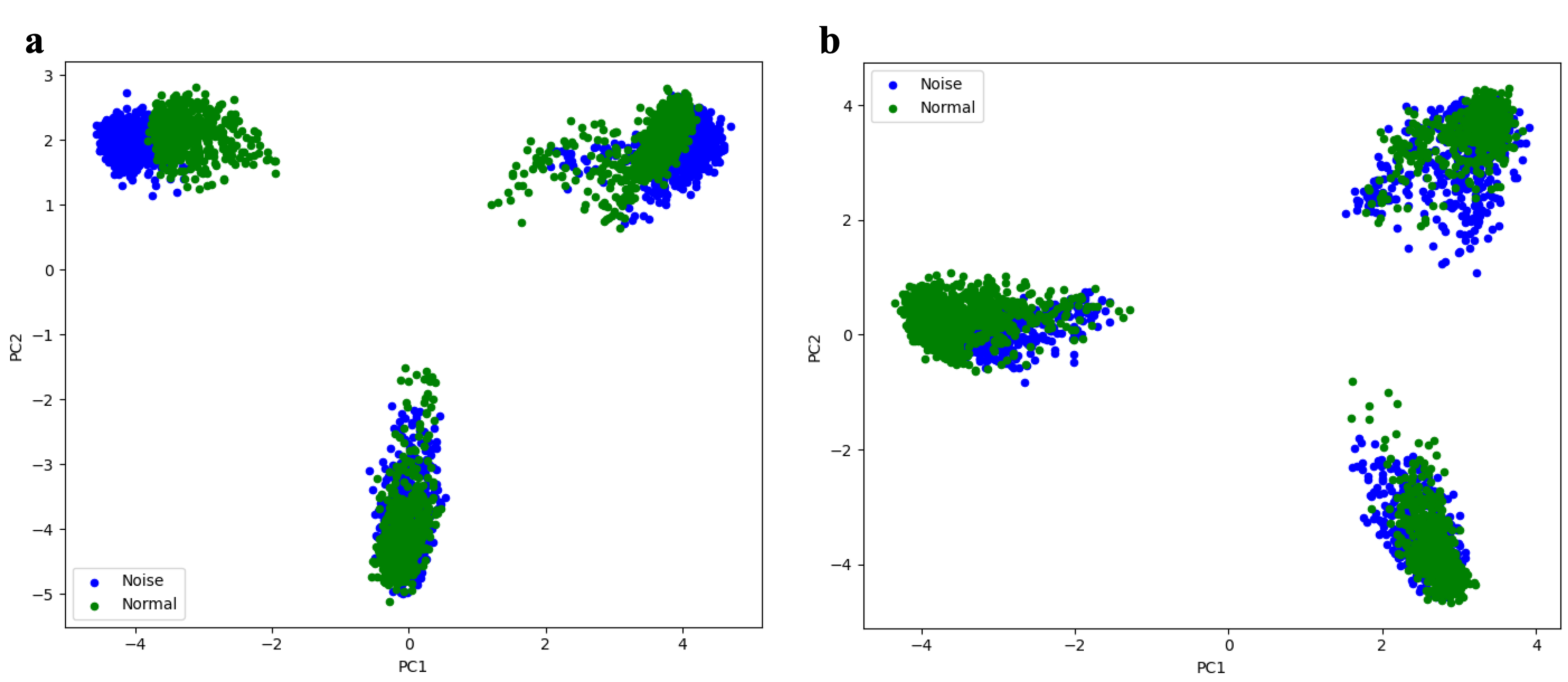}
	\caption{\textbf{PCA comparison of SAN and SAN-DWAM under noise.}Each plot shows PCA projections of hidden-layer neuronal activity during a randomly selected trajectory. Green and blue points denote normal and noisy conditions, respectively. 
	}
	\label{sup_toy}
\end{figure}

As shown in Figure~\ref{sup_toy}a, the PCA embeddings of SAN under noisy conditions (blue) deviate noticeably from those under normal conditions (green), indicating that noise substantially alters neuronal dynamics. In contrast, Figure~\ref{sup_toy}b shows that for SAN-DWAM, the noisy and noise-free data are closely aligned in the PCA space, suggesting that DWAM helps preserve stable neuronal activation despite noise. These findings demonstrate that DWAM enhances the stability and noise robustness of the network by maintaining more consistent firing patterns across conditions.

\section*{Supplementary Note 6: Additional Details on Continuous Control Experiments}
The continuous control tasks include two subtasks of varying difficulty. HalfCheetah-v3 features a 17-dimensional observation space and a 6-dimensional action space, while Ant-v3 has a 110-dimensional observation space and an 8-dimensional action space. The inclusion of tasks with different levels of complexity allows for a more comprehensive evaluation of DWAM’s performance in continuous control scenarios. In these two subtasks, the objective is to control a six-joint half-mechanical cheetah (a quadruped robot) to move forward on a plane for 1000 steps while maximizing its forward velocity. Similar to the obstacle avoidance task, we conducted experiments under various degraded conditions based on the baseline environment. These experiments include degraded input conditions, and weight uncertainty conditions.

Degraded Input: We introduced three types of disturbances to the observation data collected by the robot: “MIN”: In each trial, a random dimension of the observation space (which could represent a joint position or velocity) is replaced with its observable minimum value (-inf). “MAX”: Similar to the “MIN” condition, except that the selected dimension is replaced with its observable maximum value (inf). “GN”: In each trial, Gaussian noise sampled from $\mathcal N(0, 0.1)$ is added to every dimension of the observation space.

Weight Uncertainty: The conditions here are the same as those used in the obstacle avoidance task.

we selected two types of host SNNs: one without a homeostasis mechanism, the population-coded SAN (PopSAN)~\cite{tang2020deep}, and another with the best current homeostasis mechanism, BDETT~\cite{ding2022biologically}. Due to the similarity of the tasks, we used almost the same training setup as Ding et al.~\cite{ding2022biologically}.
\subsection*{Training Setup} 
The adopted population-coded SAN (PopSAN) is trained by using the twin-delayed deep deterministic policy gradient (TD3) off-policy algorithm~\cite{fujimoto2018addressing} and the following hyperparameter settings: $D=0.75$ for the LIF; $\eta=0.01$ and $\psi=6.0$ for the DET; $C=3.0$ for the DTT; and $\tau_s=\tau_r=1.0$ for the SRM. Compared to the settings of the obstacle avoidance tasks, the only different setting is the value of $\psi$ for the DET. Following the training protocol of the PopSAN, we set the batch size to $100$ and the learning rates to $0.0001$ for both the actor and critic networks. The reward discount factor is set to $0.99$, and the maximum length of the replay buffer is set to $1$ million.

We construct PopSAN-BioDWAM, PopSAN-DWAM, BDETT-BioDWAM, and BDETT-DWAM using the same method as the obstacle avoidance task. The BioDWAM and DWAM hyperparameters were set as follows: for the Ant-v3 environment, $\psi$ = 0.0001; for the Half Cheetah-v3 environment, $\psi$ = 0.001, $\alpha$ = 0.5, $\zeta$ = 2.0, and $n$ = 5.0. 

Notably, Ding et al. only provided the source code for the BDETT method based on the LIF model, including the network architecture and testing phase. As a result, we reproduced the BDETT implementation based on the SRM model by referring to the available LIF-based code. Moreover, due to the lack of training code, we reimplemented the training procedure following the protocol described by Ding et al.~\cite{ding2022biologically}. In certain experiments, this reproduction process may have led to lower BDETT performance compared to that reported in the original paper. However, since BDETT-DWAM is built upon BDETT and our experimental setup is designed to compare the two under identical conditions, any performance degradation would affect both methods equally. Therefore, such degradation is offset and does not affect the validity of our evaluation of DWAM.

We use PyTorch~\cite{paszke2019pytorch} to train and test all competing SNNs with an i7-14700K CPU and an NVIDIA GTX 4090 GPU.

\subsection*{Reward} 
Table~\ref{SMtab:HC_standard} and~\ref{SMtab:Ant_standard} present the success rates and homeostasis metrics of various methods in the Mujoco HalfCheetah-v3 Task and Ant-v3 Task tasks under standard scenarios, respectively. Table~\ref{SMtab:HC_Degraded} and~\ref{SMtab:Ant_Degraded} summarize the success rates and homeostasis metrics of various methods in the Mujoco HalfCheetah-v3 Task and Ant-v3 Task under different degraded conditions, respectively.

\begin{table}[ht]
\centering
\caption{Quantitative Performance of Mujoco HalfCheetah-v3 Task under standard testing condition.}
\label{SMtab:HC_standard}
\vskip 0.15in
\begin{small}
\begin{tabular}{cccc} 
\toprule
                           &                & \textbf{LIF}~       & \textbf{SRM}         \\ 
\cmidrule(lr){3-3}\cmidrule(lr){4-4}
\textbf{Type}              & \textbf{Name}  & Reward ($\sigma$)   & Reward ($\sigma$)    \\ 
\midrule
\multirow{6}{*}{Standard~} & PopSAN         & 10618 (74)          & 10824 (70)           \\
                           & PopSAN-BioDWAM & 10372 (68)          & 10827 (64)           \\
                           & PopSAN-DWAM    & \textbf{10643} (55) & \textbf{10833} (61)  \\ 
\cmidrule(lr){2-4}
                           & BDETT          & 10237 (89)          & 11312 (73)           \\
                           & BDETT-BioDWAM  & 10255 (81)          & 11306 (64)           \\
                           & BDETT-DWAM     & \textbf{10280} (65) & \textbf{11336} (90)  \\
\bottomrule
\end{tabular}
\end{small}
\vskip -0.1in
\end{table}

\begin{table}[ht]
\centering
\caption{Quantitative performance and homeostasis metrics of Mujoco HalfCheetah-v3 Tasks under various degraded conditions.}
\label{SMtab:HC_Degraded}
\vskip 0.15in
\begin{small}
\begin{tabular}{cccccccc} 
\toprule
                                                                             &                & \multicolumn{3}{c}{\textbf{LIF}~}                          & \multicolumn{3}{c}{\textbf{SRM}}                            \\ 
\cmidrule(l){3-8}
\textbf{Type}                                                                & \textbf{Name}  & Reward ($\sigma$)    & $\text{HM}_m$    & $\text{HM}_{std}$  & Reward($\sigma$)     & $\text{HM}_m$    & $\text{HM}_{std}$   \\ 
\midrule
\multirow{6}{*}{MIN~}                                                        & PopSAN         & 9993 (97)            & 0.03491          & 0.03565          & 10354 (83)           & 0.02872          & 0.02731           \\
                                                                             & PopSAN-BioDWAM & 10002 (128)          & 0.03646          & 0.03742          & 10338 (104)          & 0.02809          & 0.02670           \\
                                                                             & PopSAN-DWAM    & \textbf{10155} (81)  & \textbf{0.03205} & \textbf{0.03372} & \textbf{10413} (55)  & \textbf{0.02569} & \textbf{0.02506}  \\ 
\cmidrule(lr){2-8}
                                                                             & BDETT          & 9906 (44)            & 0.01992          & 0.02684          & 11050 (110)          & 0.01701          & 0.01774           \\
                                                                             & BDETT-BioDWAM  & 9926 (76)            & 0.02010          & 0.02775          & 11018 (70)           & 0.01689          & 0.01768           \\
                                                                             & BDETT-DWAM     & \textbf{10003} (100) & \textbf{0.01919} & \textbf{0.02678} & \textbf{11149} (87)  & \textbf{0.01633} & \textbf{0.01726}  \\ 
\midrule
\multirow{6}{*}{MAX}                                                         & PopSAN         & 10251 (86)           & 0.02823          & 0.02706          & 10511 (57)           & 0.03009          & 0.02675           \\
                                                                             & PopSAN-BioDWAM & 10186 (86)           & 0.02832          & 0.02675          & 10562 (80)           & 0.02824          & 0.02568           \\
                                                                             & PopSAN-DWAM    & \textbf{10306} (129) & \textbf{0.02721} & \textbf{0.02614} & \textbf{10639} (76)  & \textbf{0.02744} & \textbf{0.02537}  \\ 
\cmidrule(lr){2-8}
                                                                             & BDETT          & 9756 (76)            & 0.02273          & 0.02836          & 10723 (91)           & 0.01783          & 0.01763           \\
                                                                             & BDETT-BioDWAM  & 9796 (114)           & 0.02229          & 0.02806          & 10766 (73)           & 0.01770          & 0.01746           \\
                                                                             & BDETT-DWAM     & \textbf{9864} (110)  & \textbf{0.02142} & \textbf{0.02721} & \textbf{10868} (92)  & \textbf{0.01745} & \textbf{0.01731}  \\ 
\midrule
\multirow{6}{*}{GN}                                                          & PopSAN         & 4050 (80)            & 0.08455          & 0.07282          & 3641 (128)           & 0.09484          & 0.08403           \\
                                                                             & PopSAN-BioDWAM & 4006 (148)           & 0.09100          & 0.07274          & 3693 (130)           & 0.09409          & 0.08418           \\
                                                                             & PopSAN-DWAM    & \textbf{4139} (157)  & \textbf{0.08301} & \textbf{0.07180} & \textbf{3705} (147)  & \textbf{0.09314} & \textbf{0.08277}  \\ 
\cmidrule(lr){2-8}
                                                                             & BDETT          & 3425 (175)           & 0.05746          & 0.06562          & 3701 (207)           & 0.08597          & 0.07145           \\
                                                                             & BDETT-BioDWAM  & 3403 (168)           & 0.05756          & 0.06618          & 3757 (224)           & 0.08662          & 0.07184           \\
                                                                             & BDETT-DWAM     & \textbf{3498} (182)  & \textbf{0.05711} & \textbf{0.06547} & \textbf{4037} (177)  & \textbf{0.07580} & \textbf{0.06112}  \\ 
\midrule
\multirow{6}{*}{\begin{tabular}[c]{@{}c@{}}8-bit Loihi\\weight\end{tabular}} & PopSAN         & 10576 (52)           & 0.00336          & 0.00279          & 10793 (56)           & 0.00309          & 0.00281           \\
                                                                             & PopSAN-BioDWAM & 10388 (89)           & 0.00332          & 0.00301          & 10801 (97)           & 0.00354          & 0.00302           \\
                                                                             & PopSAN-DWAM    & \textbf{10651} (63)  & \textbf{0.00330} & \textbf{0.00270} & \textbf{10807} (27)  & \textbf{0.00278} & \textbf{0.00245}  \\ 
\cmidrule(lr){2-8}
                                                                             & BDETT          & 10253 (59)           & 0.00217          & 0.00241          & 11278 (128)          & 0.00501          & 0.00619           \\
                                                                             & BDETT-BioDWAM  & 10288 (102)          & 0.00328          & 0.00354          & 11247 (71)           & 0.00539          & 0.00589           \\
                                                                             & BDETT-DWAM     & \textbf{10327} (71)  & \textbf{0.00208} & \textbf{0.00231} & \textbf{11414} (83)  & \textbf{0.00444} & \textbf{0.00586}  \\ 
\midrule
\multirow{6}{*}{\begin{tabular}[c]{@{}c@{}}GN\\weight\end{tabular}}          & PopSAN         & 7158 (182)           & 0.05511          & 0.04622          & 9671 (188)           & 0.02335          & 0.02089           \\
                                                                             & PopSAN-BioDWAM & 7111 (351)           & 0.05669          & 0.04637          & 9622 (178)           & 0.02415          & 0.02130           \\
                                                                             & PopSAN-DWAM    & \textbf{7194} (342)  & \textbf{0.05511} & \textbf{0.04534} & \textbf{9882} (161)  & \textbf{0.02104} & \textbf{0.01938}  \\ 
\cmidrule(lr){2-8}
                                                                             & BDETT          & 8885 (116)           & 0.01662          & 0.01723          & 10691 (187)          & 0.01477          & 0.01324           \\
                                                                             & BDETT-BioDWAM  & 8952 (203)           & 0.01717          & 0.01710          & 10740 (124)          & 0.01500          & 0.01433           \\
                                                                             & BDETT-DWAM     & \textbf{8962} (150)  & \textbf{0.01630} & \textbf{0.01684} & \textbf{11048} (109) & \textbf{0.01387} & \textbf{0.01013}  \\ 
\midrule
\multirow{6}{*}{\begin{tabular}[c]{@{}c@{}}~30\% Zero \\weight\end{tabular}}  & PopSAN         & 4467 (999)           & 0.10544          & 0.08767          & 7255 (218)           & 0.06626          & 0.05548           \\
                                                                             & PopSAN-BioDWAM & 4327 (597)           & 0.10757          & 0.08891          & 7225 (521)           & 0.06646          & 0.05623           \\
                                                                             & PopSAN-DWAM    & \textbf{4845} (698)  & \textbf{0.10305} & \textbf{0.08743} & \textbf{7522 }(168)  & \textbf{0.06255} & \textbf{0.05270}  \\ 
\cmidrule(lr){2-8}
                                                                             & BDETT          & 1584 (555)           & 0.09059          & 0.09682          & 8167 (171)           & 0.04862          & 0.03843           \\
                                                                             & BDETT-BioDWAM  & 1425 (301)           & 0.09311          & 0.09820          & 8117 (310)           & 0.04753          & 0.04052           \\
                                                                             & BDETT-DWAM     & \textbf{1759} (347)  & \textbf{0.08837} & \textbf{0.09410} & \textbf{8219} (185)  & \textbf{0.04658} & \textbf{0.03798}  \\
\bottomrule
\end{tabular}
\end{small}
\vskip -0.1in
\end{table}

\begin{table}[ht]
\centering
\caption{Quantitative Performance of Mujoco Ant-v3 Tasks under standard testing condition.}
\label{SMtab:Ant_standard}
\vskip 0.15in
\begin{small}
\begin{tabular}{cccc} 
\toprule
                           &                & \textbf{LIF}~      & \textbf{SRM}        \\ 
\cmidrule(lr){3-3}\cmidrule(lr){4-4}
\textbf{Type}              & \textbf{Name}  & Reward ($\sigma$)  & Reward ($\sigma$)   \\ 
\midrule
\multirow{6}{*}{Standard~} & PopSAN         & 5861 (70)          & 5997 (76)           \\
                           & PopSAN-BioDWAM & 5889 (66)          & 5435 (128)          \\
                           & PopSAN-DWAM    & \textbf{5918}~(61) & \textbf{6002}~(63)  \\ 
\cmidrule(lr){2-4}
                           & BDETT          & 4736 (96)          & 5763 (58)           \\
                           & BDETT-BioDWAM  & 4735 (54)          & 5802 (65)           \\
                           & BDETT-DWAM     & \textbf{4761}~(63) & \textbf{5885}~(84)  \\
\bottomrule
\end{tabular}
\end{small}
\vskip -0.1in
\end{table}

\begin{table}[ht]
\centering
\caption{Quantitative performance and homeostasis metrics of Mujoco Ant-v3 Tasks under various degraded conditions.}
\label{SMtab:Ant_Degraded}
\vskip 0.15in
\begin{small}
\begin{tabular}{cccccccc} 
\toprule
                                                                             &                & \multicolumn{3}{c}{\textbf{LIF}~}                         & \multicolumn{3}{c}{\textbf{SRM}}                           \\ 
\cmidrule(l){3-8}
\textbf{Type}                                                                & \textbf{Name}  & Reward ($\sigma$)   & $\text{HM}_m$    & $\text{HM}_{std}$  & Reward($\sigma$)    & $\text{HM}_m$    & $\text{HM}_{std}$   \\ 
\midrule
\multirow{6}{*}{MIN~}                                                        & PopSAN         & 3521 (550)          & 0.09214          & 0.06623          & 5687 (82)           & 0.05702          & 0.06007           \\
                                                                             & PopSAN-BioDWAM & 3539 (216)          & 0.08624          & 0.07033          & 4967 (57)           & 0.06404          & 0.08363           \\
                                                                             & PopSAN-DWAM    & \textbf{3794} (141) & \textbf{0.06646} & \textbf{0.06037} & \textbf{5709} (60)  & \textbf{0.05534} & \textbf{0.05897}  \\ 
\cmidrule(lr){2-8}
                                                                             & BDETT          & 4186 (97)           & 0.03949          & 0.06111          & 4275 (198)          & 0.05800          & 0.05183           \\
                                                                             & BDETT-BioDWAM  & 4033 (113)          & 0.04830          & 0.06803          & 4272 (166)          & 0.05973          & 0.06091           \\
                                                                             & BDETT-DWAM     & \textbf{4214} (113) & \textbf{0.03926} & \textbf{0.06099} & \textbf{4388} (162) & \textbf{0.05622} & \textbf{0.05048}  \\ 
\midrule
\multirow{6}{*}{MAX}                                                         & PopSAN         & 4795 (71)           & 0.07907          & 0.07715          & 5430 (101)          & 0.05057          & 0.04851           \\
                                                                             & PopSAN-BioDWAM & 4808 (104)          & 0.07942          & 0.07685          & 4369 (176)          & 0.06989          & 0.07343           \\
                                                                             & PopSAN-DWAM    & \textbf{4900} (95)  & \textbf{0.07875} & \textbf{0.07641} & \textbf{5581} (118) & \textbf{0.04825} & \textbf{0.04737}  \\ 
\cmidrule(lr){2-8}
                                                                             & BDETT          & 3638 (70)           & 0.04927          & 0.07121          & 4510 (79)           & 0.06381          & 0.06269           \\
                                                                             & BDETT-BioDWAM  & 2779 (91)           & 0.08293          & 0.07792          & 4465 (90)           & 0.06342          & 0.06264           \\
                                                                             & BDETT-DWAM     & \textbf{3691} (92)  & \textbf{0.04878} & \textbf{0.07057} & \textbf{4516} (69)  & \textbf{0.06299} & \textbf{0.06224}  \\ 
\midrule
\multirow{6}{*}{GN}                                                          & PopSAN         & 2160 (336)          & 0.12501          & 0.09122          & 2109 (96)           & 0.09279          & 0.08004           \\
                                                                             & PopSAN-BioDWAM & 2216 (255)          & 0.11825          & 0.09035          & 2088 (107)          & 0.10151          & 0.10768           \\
                                                                             & PopSAN-DWAM    & \textbf{2541} (214) & \textbf{0.10536} & \textbf{0.08702} & \textbf{2153} (111) & \textbf{0.09187} & \textbf{0.07959}  \\ 
\cmidrule(lr){2-8}
                                                                             & BDETT          & 2182 (129)          & 0.07004          & 0.09268          & 2275 (101)          & 0.09007          & 0.07314           \\
                                                                             & BDETT-BioDWAM  & 1714 (83)           & 0.09131          & 0.09800          & 2045 (90)           & 0.09975          & 0.08309           \\
                                                                             & BDETT-DWAM     & \textbf{2230} (130) & \textbf{0.06949} & \textbf{0.09205} & \textbf{2517} (176) & \textbf{0.08012} & \textbf{0.06231}  \\ 
\midrule
\multirow{6}{*}{\begin{tabular}[c]{@{}c@{}}8-bit Loihi\\weight\end{tabular}} & PopSAN         & 5904 (43)           & 0.00827          & 0.01115          & 6044 (74)           & 0.00720          & 0.00672           \\
                                                                             & PopSAN-BioDWAM & 5936 (51)           & 0.00799          & 0.01023          & 5452 (100)          & 0.00682          & 0.00613           \\
                                                                             & PopSAN-DWAM    & \textbf{5948} (53)  & \textbf{0.00724} & \textbf{0.00953} & \textbf{6050} (56)  & \textbf{0.00574} & \textbf{0.00501}  \\ 
\cmidrule(lr){2-8}
                                                                             & BDETT          & 4732 (113)          & 0.00290          & 0.00401          & 5732 (66)           & 0.00786          & 0.00958           \\
                                                                             & BDETT-BioDWAM  & 4699 (69)           & 0.00470          & 0.00466          & 5727 (64)           & 0.00806          & 0.00930           \\
                                                                             & BDETT-DWAM     & \textbf{4774} (60)  & \textbf{0.00257} & \textbf{0.00337} & \textbf{5822} (23)  & \textbf{0.00715} & \textbf{0.00908}  \\ 
\midrule
\multirow{6}{*}{\begin{tabular}[c]{@{}c@{}}GN\\weight\end{tabular}}          & PopSAN         & 1324 (862)          & 0.18511          & 0.11284          & 5580 (92)           & 0.02456          & 0.02102           \\
                                                                             & PopSAN-BioDWAM & 1420 (538)          & 0.16646          & 0.11603          & 5017 (97)           & 0.02663          & 0.02395           \\
                                                                             & PopSAN-DWAM    & \textbf{1519} (700) & \textbf{0.14485} & \textbf{0.09669} & \textbf{5627} (104) & \textbf{0.02150} & \textbf{0.01941}  \\ 
\cmidrule(lr){2-8}
                                                                             & BDETT          & 3626 (251)          & 0.02328          & 0.02773          & 4939 (109)          & 0.03999          & 0.03502           \\
                                                                             & BDETT-BioDWAM  & 2905 (87)           & 0.03265          & 0.03145          & 5550 (73)           & 0.02810          & 0.02518           \\
                                                                             & BDETT-DWAM     & \textbf{3637} (280) & \textbf{0.02216} & \textbf{0.02762} & \textbf{5558} (96)  & \textbf{0.01728} & \textbf{0.01520}  \\ 
\midrule
\multirow{6}{*}{\begin{tabular}[c]{@{}c@{}}~30\% Zero \\weight\end{tabular}}  & PopSAN         & 1580 (653)          & 0.12429          & 0.10770          & 4091 (258)          & 0.07828          & 0.06308           \\
                                                                             & PopSAN-BioDWAM & 1703 (775)          & 0.13470          & 0.10780          & 3884 (312)          & 0.08358          & 0.07115           \\
                                                                             & PopSAN-DWAM    & \textbf{1873} (764) & \textbf{0.12242} & \textbf{0.10198} & \textbf{4219} (435) & \textbf{0.07659} & \textbf{0.06250}  \\ 
\cmidrule(lr){2-8}
                                                                             & BDETT          & 1937 (566)          & 0.05455          & 0.07207          & 4949 (106)          & 0.03999          & 0.03502           \\
                                                                             & BDETT-BioDWAM  & 1313 (490)          & 0.06653          & 0.08142          & 4890 (163)          & 0.04159          & 0.03685           \\
                                                                             & BDETT-DWAM     & \textbf{2157} (504) & \textbf{0.05336} & \textbf{0.07075} & \textbf{5030} (150) & \textbf{0.03711} & \textbf{0.03292}  \\
\bottomrule
\end{tabular}
\end{small}
\vskip -0.1in
\end{table}